\documentclass[11pt]{scaleai-paper}

\usepackage{amsmath}
\usepackage{amsfonts}
\usepackage{amssymb}
\usepackage{amsthm}
\usepackage{booktabs}
\usepackage{tabularx}
\usepackage{tabulary}
\usepackage{multirow}
\usepackage{subcaption}
\usepackage{float}
\usepackage[square,numbers,sort&compress]{natbib}
\usepackage{xspace}
\usepackage{url}
\usepackage[colorlinks=true,linkcolor=scaleLink,citecolor=scaleLink,urlcolor=scaleLink]{hyperref}
\usepackage[capitalise,nameinlink]{cleveref}
\usepackage[utf8]{inputenc} % allow utf-8 input
\usepackage[T1]{fontenc}    % use 8-bit T1 fonts
\usepackage{hyperref}       % hyperlinks
\usepackage{url}            % simple URL typesetting
\usepackage{booktabs}       % professional-quality tables
\usepackage{amsfonts}       % blackboard math symbols
\usepackage{nicefrac}       % compact symbols for 1/2, etc.
\usepackage{xcolor}         % colors
\usepackage{microtype}
\usepackage{dirtytalk}
\usepackage{graphicx}
\usepackage{amsmath}
\usepackage{float}
\usepackage{amssymb}
\usepackage{tabularx}
\usepackage{booktabs}
\usepackage{ltablex} % Allows tables to span multiple pages
\usepackage{algorithm}
\usepackage{algorithmic}
\usepackage{multirow}
\usepackage[table]{xcolor} % For row shading
\usepackage[normalem]{ulem}
\newcommand{\cmark}{\checkmark}
\newcommand{\xmark}{\scalebox{0.85}{$\times$}}
\definecolor{recoverygray}{gray}{0.92}
\definecolor{tableblue}{RGB}{215, 243, 250}

\definecolor{lightgray}{gray}{0.95}
\usepackage{amsmath,amsfonts,bm}

\def\eqref#1{equation~\ref{#1}}
\def\1{\bm{1}}

\DeclareMathAlphabet{\mathsfit}{\encodingdefault}{\sfdefault}{m}{sl}
\SetMathAlphabet{\mathsfit}{bold}{\encodingdefault}{\sfdefault}{bx}{n}

\newcolumntype{Y}{>{\RaggedRight\arraybackslash}X}
\let\svthefootnote\thefootnote
\newcommand\freefootnote[1]{%
  \let\thefootnote\relax%
  \footnotetext{#1}%
  \let\thefootnote\svthefootnote%
}

\contact{\texttt{research@scale.com}}

\title{Insights Generator: Systematic Corpus-Level Trace Diagnostics for LLM Agents}

\author[$\dagger$]{Akshay Manglik}
\author[$*$]{Vijay S. Kalmath}
\author{Jason Qin}
\author{Apaar Shanker}
\author[$\dagger$]{Kaustubh Deshpande}
\author{Yash Maurya}
\author{Veronica Chatrath}
\author{Levi Lentz}
\author{Yuan (Emily) Xue}

\affil{Scale AI}

\begin{document}
\maketitle
\freefootnote{${}^\dagger$Work conducted while at Scale AI.}
\freefootnote{${}^*$Corresponding author: \texttt{vijay.kalmath@scale.com}}

\begin{abstract}
Diagnosing failures in LLM agents remains largely manual. Practitioners inspect a small subset of execution traces, form ad-hoc hypotheses, and iterate. This process misses patterns that only emerge across trace populations and does not scale to production corpora where individual traces span tens of thousands of tokens. We formalize the problem of corpus-level trace diagnostics. Given a corpus of execution traces, the goal is to produce grounded natural-language insights that characterize systematic behavioral patterns across trace groups, each linked to supporting evidence. We present the \textbf{Insights Generator (IG)}, a multi-agent system that answers diagnostic questions by proposing and testing hypotheses across the trace corpus to produce an evidence-backed insights report. We evaluate IG across qualitative and objective dimensions, spanning rubric-based report assessment and downstream performance improvements achieved by implementing IG insights. Human experts using IG reports improve scaffold performance by 30.4\,pp over the unmodified baseline scaffold, and coding agents leveraging IG-derived insights show consistent and stable gains. Across benchmarks, IG's \textit{scout-investigator} architecture produces findings comparable in detection coverage to competing approaches, while domain experts rated IG reports as leading depth and evidence quality.
\end{abstract}

\section{Introduction}

Agents have become central to modern research and technology, from software-engineering and expert-reasoning benchmarks~\citep{swebench, hle} to RL environments~\citep{swegym} and automated agent design~\citep{ADAS, vero}. These agentic systems produce long execution traces of reasoning steps, tool calls, and environment interactions. Debugging typically starts from evaluation results that flag failed runs; practitioners then inspect a few traces and form hypotheses~\citep{TRAIL_paper, wink}. Evaluations are defined over final task outcomes, so they yield aggregate signals but reveal neither intermediate behaviors nor patterns across the broader space of scenarios, and many errors, inefficiencies, and performance differences stay hard to identify~\citep{TRAIL_paper, demistifying}. This is especially true for failures localized to specific subsets of inputs, task types, or execution contexts that are not individually prominent enough to noticeably shift aggregate metrics. Such patterns are invisible to standard evaluation unless the practitioner already knows to look for them.

Many of these signals emerge only at the cohort level. Systematic differences in tool usage, recurring reasoning structures, and patterns correlated with strong or weak performance become visible only when traces are analyzed collectively. Cohort comparison is a particularly powerful form of this analysis: examining groups of traces that vary along a single controlled axis, such as model variant, prompt version, or task category, isolates the behavioral differences attributable to that axis and provides a stronger targeted diagnostic signal than aggregate metrics alone. Beyond diagnosis of known failure modes, there is a distinct need for open-ended \emph{discovery}. Silent failures, behavioral trends that reduce reliability without triggering explicit error signals, are a key example; they rarely appear in evaluation dashboards and require population-level analysis to detect. We refer to such findings as \textbf{trace insights}: grounded natural-language statements characterizing salient patterns across groups of traces, each supported by trace-level evidence. \label{sec:intro_example}For example, a representative IG insight on HLE might read: \texttt{84\% of incorrect Python-using traces exhibit `silent computation failures': the code executes without errors but implements the wrong mathematical model. Cross-checking fails to catch these because it verifies internal consistency within the wrong framework rather than challenging the framework itself. Observed in 94/112 incorrect Python-using traces (37.6\% of all traces); 0/19 cross-checking traces detected the modeling error.}

Surfacing trace insights at corpus scale presents two challenges. \textbf{Scale and Complexity:} Trace corpora, the collections of traces gathered across many runs of an agent, routinely exceed any single model context window even when each individual trace fits within it. Individual traces interleave thousands of tokens of reasoning, tool calls, and environment state. The patterns of interest (silent failures, systematic inefficiencies, behavioral differences between cohorts) are semantic and not reducible to keyword matching or predefined taxonomies~\citep{oolong, holobench}. Prior work addresses fragments of this problem through single-trajectory debugging and failure attribution~\citep{agentdebug2025, dover, agentrx, TRAIL_paper, whoandwhen}, classification under fixed failure taxonomies~\citep{multiAgentSystems, demistifying}, outcome-conditioned distillation of lessons into skills~\citep{trace2skill}, harness optimization in which a proposer reads the raw traces of prior candidates~\citep{metaharness}, or recursive search over trace snippets~\citep{halo}. Proposing descriptions on samples and validating them at scale has precedent for datasets~\citep{d5, vibecheck}, but to our knowledge no trace-analysis system separates hypothesis proposal on sampled traces from statistical validation over the full corpus (Section~\ref{sec:related_work}). \textbf{No Evaluation Standard:} Trace insights are free-form claims about population-level behavior; whether such a claim is correct, grounded in cited evidence, and useful to a downstream practitioner cannot be reduced to a single existing metric. Existing benchmarks score per-trajectory failure attribution against annotated or injected failures~\citep{TRAIL_paper, whoandwhen, whowhenpro} or insights against ground truth planted in the data~\citep{insightbench}, and none measures a trace-analysis report by the improvement it enables.

We present \emph{Insights Generator (IG)}, which addresses the first challenge through iterative hypothesis-driven aggregation: it decomposes an open-ended diagnostic question into testable hypotheses, validates each at corpus scale, and synthesizes evidence-backed findings. Our contributions are:
\begin{itemize}
    \item \textbf{Iterative corpus-level trace analysis.} IG separates hypothesis generation from corpus-scale validation and routes all trace access and processing through a stateful Python layer, enabling a decompose-hypothesize-validate loop that operates over corpora exceeding any single context window. This surfaces patterns, such as the silent computation failures above, that single-pass and per-trajectory methods miss (Section~\ref{sec:related_work}).
    \item \textbf{Evaluation framework for insights.} We introduce a four-setting framework that varies who evaluates (rubric-based LLM judge vs. human expert) and what is measured (report quality vs. downstream scaffold impact), and apply it to a range of trace-analysis systems on multiple benchmarks. IG ranks first or ties for first in every setting, and the pairwise rankings agree across two LLM judge families (Spearman 0.77 to 0.99). To our knowledge it is the first evaluation that measures trace-analysis reports by the agent improvement they enable (Section~\ref{sec:evaluation}).
    \item \textbf{Practitioner and automated impact.} Subject-matter experts using IG reports improve scaffold performance by 30.4\,pp over the unmodified baseline, nearly doubling the 16.2\,pp gain from the strongest alternative system ($p=0.016$). This gap is consistent with IG's higher scores on evidence depth and mechanism explanation in the automated evaluation (Section~\ref{sec:he_intervention}). In an automated patching loop on APEX-Agents, IG reports produce the largest held-out gain of any report source, and IG report quality rises with analysis budget: citation validity and judged quality increase across the capped budget tiers under both judge families (Sections~\ref{sec:patcher} and~\ref{sec:budget}).
\end{itemize}

\begin{figure}[t]
\begin{center}
\centerline{\includegraphics[width=\columnwidth]{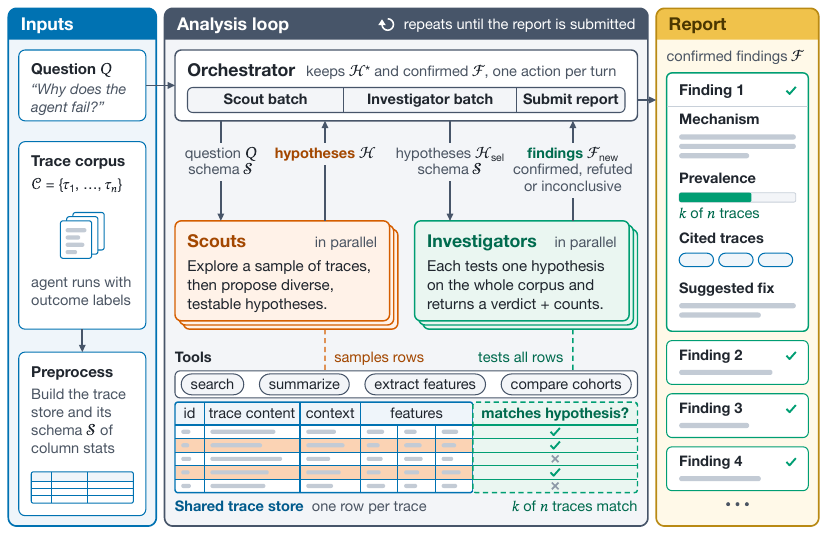}}
\caption{IG system overview. Scouts propose hypotheses from a small sample of traces (amber rows), and the Orchestrator prioritizes them. Each Investigator tests one hypothesis against every trace, computing a feature for all $n$ traces (teal column) and comparing it across outcomes. Only confirmed findings enter the report, each with a prevalence, cited traces, and a fix. Algorithm~\ref{alg:ig_loop} formalizes the loop.}
\label{fig:ig_overview}
\end{center}
\vspace{-10mm}
\end{figure}

\section{Related Work}
\label{sec:related_work}

\paragraph{Single-Trajectory Error Analysis.} Most agent debugging work analyzes one trajectory at a time, often by counterfactual replay: AgentDebug~\citep{agentdebug2025} reasons about alternative decision points, AgenTracer~\citep{agentracer} builds failure-attribution data through counterfactual replay and programmatic fault injection and trains a lightweight attribution model, and DoVer~\citep{dover} replays trajectories with targeted interventions. AgentRx~\citep{agentrx} infers constraints from tool schemas and checks whether agent actions satisfy them. Taxonomy-based methods (MAST~\citep{multiAgentSystems}, AgentFail~\citep{demistifying}) classify known failure types but cannot surface modes the taxonomy did not anticipate, and failure-attribution benchmarks such as Who\&When~\citep{whoandwhen} ask which agent and which step caused a given failure. Per-trace diagnosis remains hard: on TRAIL~\citep{TRAIL_paper} the best models score only 11\%. Some, such as DoVer, already pair failure hypotheses with active tests, but by intervention on one trajectory at a time. These trajectory-focused methods excel at individual remediation, but miss systemic patterns that only emerge across corpora of agent runs, which is the gap our corpus-level approach addresses.

\paragraph{Agentic Insight Generation.} More recent work has leveraged agent-based decompositions to surface insights from agent execution traces. Trace2Skill~\citep{trace2skill} dispatches a flat parallel fleet of analyst subagents over individual trajectories to identify skill patches, and then performs hierarchical consolidation to merge individual patches into a conflict-free skill directory consumed by the agent at inference time. HALO~\citep{halo} employs a Recursive Language Model~\citep{rlm} that analyzes traces through tools like search, view, count, and code, and recursively invokes itself on the relevant snippets, producing a diagnostic report handed to a downstream coding agent for harness fixes. Docent~\citep{docent} uses language models to summarize, cluster, and search agent transcripts for human review. Closest in method, D5~\citep{d5} proposes hypotheses about how two text corpora differ from samples and validates them across the corpora, and VibeCheck~\citep{vibecheck} applies the same pattern to qualitative differences between LLMs, scoring each proposed trait with LLM judges. IG builds on this pattern and differs from prior trace-analysis systems in two ways. Its typed Scout and Investigator roles, rather than a flat or recursive decomposition, separate pattern proposal on samples from statistical validation against every trace, so each finding carries a measured prevalence over a named cohort; and all trace access runs through a stateful Python data layer, so findings reflect corpus-level distributions rather than per-trajectory observations.

\paragraph{Agent Optimization Frameworks.} Several recent works frame agent improvement as an automated optimization task. VeRO~\citep{vero} introduces an evaluation harness for \emph{agent optimization}: a coding agent iteratively modifies a target agent's implementation, prompts, tools, and orchestration logic, and is evaluated on the lift it achieves against a fixed benchmark. AFlow~\citep{aflow} and ADAS~\citep{ADAS} similarly explore automated search over agent-as-code representations. Meta-Harness~\citep{metaharness} delegates both diagnosis and proposal to a single coding-agent proposer that reads the source code, scores, and execution traces of all prior candidates through a filesystem, rather than externalizing the improvement logic in a hand-designed search loop. These systems treat the agent as an artifact to be improved against a specific evaluation set. Our work is complementary: rather than optimizing an agent toward a fixed target, IG surfaces behavioral insights latent in a trace corpus, including patterns that no predefined objective would have targeted. These insights can then be consumed by an agent-optimization framework in a closed loop.

\section{Insights Generator}
\label{sec:ig}

\paragraph{Overview.} Insights Generator (IG) is a multi-agent analysis system for discovering and validating behavioral patterns across large trace corpora without requiring full-context loading of raw trajectories. The system is motivated by a practical bottleneck in agent debugging: manual inspection of small numbers of traces may be sufficient to form hypotheses, but is insufficient to validate them at scale. IG addresses this by separating hypothesis discovery from hypothesis validation, and by routing all analysis through a Python data processing layer rather than passing raw trace content to LLM context directly. Agents interact with traces exclusively through analysis primitives (summarization, extraction, cohort comparison), called from a single code-execution tool, that return only aggregated results to model context. Figure~\ref{fig:ig_overview} illustrates the system; Algorithm~\ref{alg:ig_loop} (Appendix~\ref{app:algorithm}) formalizes the multi-round loop.

\paragraph{Task definition.} The input is a diagnostic question $Q$ and a corpus of agent execution traces $\mathcal{C} = \{\tau_1, \dots, \tau_n\}$, each interleaving inputs, reasoning steps, tool calls, environment observations, and outputs, optionally with metadata such as task-specific model inputs, outcome labels, configuration variants, or timestamps. The output is a set of corpus-level insights describing salient behavioral patterns across subsets of $\mathcal{C}$, each supported by trace-level evidence. Unlike skill distillation~\citep{trace2skill} or harness optimization~\citep{metaharness}, the output is itself the object of evaluation: each insight states a prevalence over a named cohort and cites trace IDs, so both can be checked deterministically against the corpus. With labels, insights can characterize systematic differences between cohorts (e.g., success versus failure); without them, the task reduces to unsupervised pattern discovery.

\paragraph{Agent roles.} The \textbf{Orchestrator} is the sole user-facing agent. It decomposes the objective, dispatches subagents in parallel, judges whether findings suffice across rounds, and synthesizes the final report, inspecting traces minimally to keep its context free for coordination. A \textbf{Scout} explores a representative trace sample with LLM summarization and extraction tools and proposes diverse testable hypotheses; its objective is breadth, not validation. An \textbf{Investigator} receives one hypothesis, validates it at corpus scale with LLM extraction and programmatic analysis (distributional statistics, cohort comparisons), and returns a finding labeled confirmed, refuted, or inconclusive with quantitative evidence. Prompts are in Appendix~\ref{app:agent_prompts}; models and runtime settings are in Appendix~\ref{app:implementation_details}.

\paragraph{Data layer.} Agents operate over a processed trace store rather than raw text, so they can apply arbitrary data-science transformations in Python. Traces and extracted features share a short string identifier rather than a UUID, which reduces token usage; consistent with prior observations that LLMs mis-copy long random identifiers and that copying fidelity degrades with identifier length~\citep{hsieh2024ruler, dai2024deniahl}, we also observe fewer hallucinated references with short identifiers. A schema cache $\mathcal{S}$ of per-column statistics, null rates, value distributions, and notable correlations is computed automatically and injected into every subagent's context at dispatch. A hybrid semantic-keyword vector store over the traces serves content retrieval to all agents as a tool. Section~\ref{sec:budget} reports how removing components of this layer affects downstream scaffold improvement.

\paragraph{Programmatic tool use.} Each subagent works in a stateful Python environment with IG's analysis primitives pre-injected as functions; to the agent, the interface is a single code-execution tool. The primitives cover \emph{trace inspection} (loading traces, retrieving raw content, querying the schema cache, hybrid search), \emph{LLM-driven analysis} (structured feature extraction, per-trace and group-level summarization, A/B cohort comparison), and \emph{data management} (persisting computed columns and finding-supporting cohorts, refreshing the cached DataFrame, consolidating extraction parquets). The Orchestrator keeps a coordination subset, the Scout adds extraction to that subset, and the Investigator further adds cohort comparison and finding persistence. Agents chain operations in one code block, so intermediate trace content never enters model context. Appendix~\ref{app:tools} lists every primitive and its per-role availability.

\paragraph{Iterative analysis loop.} At each turn the Orchestrator inspects the candidate hypotheses $\mathcal{H}^*$ and confirmed findings $\mathcal{F}$ and takes one of three actions: dispatch a parallel batch of Scouts to surface new hypotheses, dispatch a parallel batch of Investigators to validate prioritized hypotheses at corpus scale, or submit the final report. Scout and Investigator rounds can interleave in any order and number, depending on coverage. When the Orchestrator judges that no more information is needed, it synthesizes the confirmed findings from all rounds into the report. Algorithm~\ref{alg:ig_loop} (Appendix~\ref{app:algorithm}) formalizes the loop.

\section{Evaluation}
\label{sec:evaluation}

\begin{figure}[t]
\begin{center}
\centerline{\includegraphics[width=0.9\columnwidth]{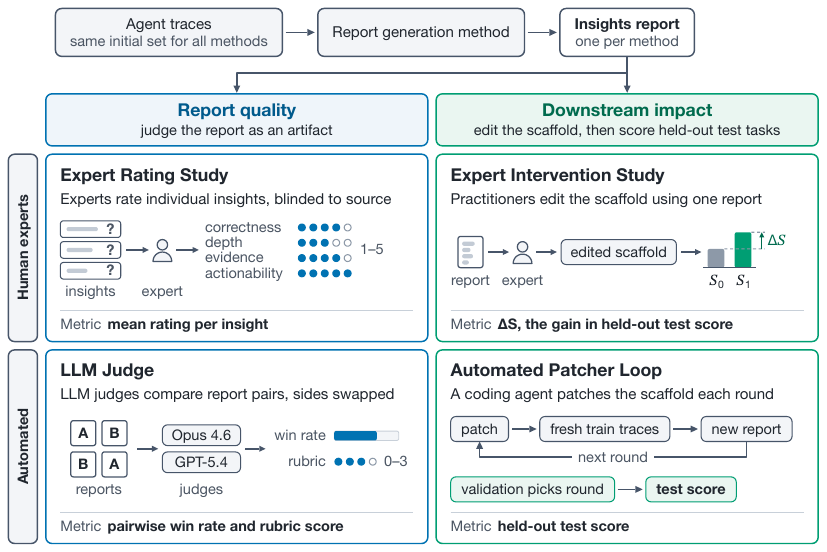}}
\caption{The four evaluation settings, by who evaluates (rows) and what is measured (columns). Every setting receives a report generated from the same training traces, so only the report source varies. The patcher loop regenerates the report each round and is scored on held-out test tasks.}
\label{fig:evaluation_overview}

\end{center}
\end{figure}

We evaluate IG along two axes, \emph{who evaluates} (an LLM judge or human experts) and \emph{what is measured} (report quality or downstream scaffold impact), which gives four settings (Figure~\ref{fig:evaluation_overview}): the LLM judge (Section~\ref{sec:llm_judge}); the \textbf{Expert Intervention Study}, in which experts modify a scaffold given a report (Section~\ref{sec:he_intervention}); the \textbf{Expert Rating Study}, in which experts score individual insights (Section~\ref{sec:he_judge}); and the automated patcher loop (Section~\ref{sec:patcher}). Because IG is meant to support agent developers rather than replace them, the Expert Intervention Study is our primary measure of downstream utility; the patcher loop is a complementary test of end-to-end insight transfer. Section~\ref{sec:budget} reports component ablations, cost scaling, and backbone choice.

We use three benchmarks (Appendix~\ref{app:benchmark_details}): \textbf{SSB} (SpreadsheetBench)~\citep{spreadsheetbench, ssbverified}, spreadsheet-manipulation tasks with a deterministic cell-match grader and the primary benchmark for the human studies; \textbf{HLE} (Humanity's Last Exam)~\citep{hle}, expert-level academic questions with an LLM grader; and \textbf{APEX-Agents}~\citep{apexagents}, long-horizon, document-grounded professional tasks with roughly 13$\times$ more tool calls per trace than SSB, which serves as the harder second domain for the judge, intervention, and patcher settings.

\subsection{Report Quality under an LLM Judge}
\label{sec:llm_judge}
We compare IG with four systems given the same corpora and the same canonical diagnostic question: \textbf{Single-Agent Coding} (Claude Code, \citealp{claudecode}, with its native tool surface); \textbf{CC Subagents} (Claude Code with subagents), which adds subagent dispatch and IG's scout-investigator split without IG's data layer, so each subagent works in an isolated context and returns a short summary; a \textbf{Trace2Skill}-style system~\citep{trace2skill} with flat parallel analysts and hierarchical consolidation; and \textbf{RLM}~\citep{rlm}, a vanilla Recursive Language Model at maximum depth~2. Together these vary tool surface, multi-agent topology, and role decomposition. A shared module normalizes all reports to one JSON schema, and every SSB and HLE metric is a mean over three report replicates per system, with one report per system on APEX-Agents (Appendix~\ref{app:baseline_details}). Two further comparisons use the same corpora. Four statistical baselines apply fault localization to trace elements~\citep{liblit2005} and significance tests, failure subgroups, and clustering to 40--60 hand-derived trace features, and write up either those outputs or the traces they retrieve in a single Claude Opus~4.6 call (Appendix~\ref{app:stat_baselines}). IG and all four systems also run on two open-weight backbones, DeepSeek V4 Flash and GLM-5.3 Flash (Appendix~\ref{app:open_models}).

\begin{table}[t]
\centering
\caption{LLM-judge results (Claude Opus~4.6) on SSB, HLE, and APEX-Agents. \textbf{Pairwise WR}: position-swapped round-robin win rate (\%); Avg is over SSB and HLE only. \textbf{Quality}: 0--3 rubric scores for mechanism (Mech.), specificity (Spec.), actionability (Act.), and their mean (Comp.), averaged over SSB, HLE, and APEX-Agents. Tournament pools and the one-report APEX-Agents protocol: Appendix~\ref{app:llm_judge_details}. IG across backbones: win rate against the four Opus~4.6 baselines, one report each, so IG on Opus~4.6 scores differently there than in the main block. Bold: best per column in the main block.}
\label{tab:llm_judge_combined}
\vspace{4pt}
\small
\setlength{\tabcolsep}{4.5pt}
\begin{tabular}{l cccc cccc}
\toprule
& \multicolumn{4}{c}{Pairwise WR (\%)} & \multicolumn{4}{c}{Quality (0--3)} \\
\cmidrule(lr){2-5} \cmidrule(lr){6-9}
System & SSB & HLE & Avg & APEX & Mech. & Spec. & Act. & Comp. \\
\midrule
IG                  & \textbf{70.3} & \textbf{81.8} & \textbf{76.0} & \textbf{88.9} & \textbf{1.48} & \textbf{1.42} & \textbf{1.32} & \textbf{1.41} \\
CC Subagents        & 61.5 & 59.9 & 60.7 & 77.8 & 1.17 & 1.14 & 1.12 & 1.14 \\
Trace2Skill         & 54.7 & 47.9 & 51.3 & 61.1 & 1.11 & 1.08 & 1.12 & 1.10 \\
RLM                 & 12.5 &  8.9 & 10.7 & 22.2 & 1.06 & 1.06 & 0.90 & 1.01 \\
Single-Agent Coding & 16.7 & 12.5 & 14.6 &  0.0 & 0.67 & 0.68 & 0.65 & 0.67 \\
Statistics + LLM write-up & 0.0 & 0.0 & 0.0 & 10.0 & 0.61 & 0.55 & 0.65 & 0.60 \\
\midrule
\multicolumn{9}{l}{\textit{IG architecture ablations}} \\
IG, Generic Subagents & 57.3 & 74.0 & 65.6 & 72.2 & 1.28 & 1.22 & 1.12 & 1.21 \\
IG, Orchestrator-only & 31.8 & 28.1 & 29.9 & 38.9 & 0.85 & 0.84 & 0.78 & 0.82 \\
\midrule
\multicolumn{9}{l}{\textit{IG across backbones}} \\
IG, Opus~4.6 & 75.0 & 87.5 & 81.3 & 75.0 & 1.51 & 1.40 & 1.29 & 1.40 \\
IG, GLM-5.3 Flash & 100.0 & 100.0 & 100.0 & 100.0 & 1.78 & 1.77 & 1.66 & 1.74 \\
IG, DeepSeek V4 Flash & 50.0 & 87.5 & 68.8 & 75.0 & 1.05 & 1.04 & 0.91 & 1.00 \\
\bottomrule
\end{tabular}
\end{table}

\paragraph{Protocol.} The judge pools all systems' findings and clusters them into a \emph{union-gold} set of canonical observations. It scores each report on \textbf{coverage}, the share of clusters for which the report cites at least one of the cluster's traces, and on a 0--3 rubric per dimension against the clusters (Appendix~\ref{app:llm_judge_prompt}); Table~\ref{tab:llm_judge_combined} reports mechanism, specificity, and actionability. Because the union-gold reference is built from the evaluated systems' own pooled findings, coverage is not an independent measure, so we omit it from Table~\ref{tab:llm_judge_combined} and report it in Appendix~\ref{app:llm_judge_details}. Separately, a \textbf{pairwise} round-robin tournament asks the judge to prefer one of two reports or declare a tie, and computes win rate as $(\text{wins} + 0.5 \times \text{ties}) / n_{\text{rounds}}$ with no gold reference. Because the judge shares its model (Claude Opus~4.6) with every analysis system, we also re-ran the full pipeline with GPT-5.4 as the judge on the same reports and rubric, as a control for judge-family bias; the two judges' rankings agree at Spearman $\rho = 0.99$ (SSB), $0.95$ (HLE), and $0.77$ (APEX-Agents) (Table~\ref{tab:cross_judge}, Appendix~\ref{app:llm_judge_details}).

\paragraph{Results.} IG ranks first on all three benchmarks, with an SSB and HLE average pairwise win rate of 76.0\% against 60.7\% for the next-best system (CC Subagents) and 88.9\% against 77.8\% on APEX-Agents, and leads on every rubric dimension (composite 1.41 vs.\ 1.14 for CC Subagents; Table~\ref{tab:llm_judge_combined}). The four statistical baselines win 0 of 48 rounds against IG, and the strongest scores a rubric composite of 0.60 (Table~\ref{tab:llm_judge_combined}, Appendix~\ref{app:stat_baselines}). To see where this lead comes from, we vary IG along two axes, its architecture and its backbone. On the architecture axis, the ablations locate the gain in two steps: dispatching subagents raises the average win rate from 29.9\% (\textbf{Orchestrator-only}) to 65.6\% (\textbf{Generic Subagents}, one untyped subagent role), and typing them as Scouts and Investigators raises it to 76.0\% (full IG), with IG beating Generic Subagents in 14 of 18 SSB rounds; the ordering holds in pairwise win rate under GPT-5.4 (Table~\ref{tab:cross_judge}), per-benchmark breakdowns are in Appendix~\ref{app:llm_judge_details}, and downstream component ablations in Section~\ref{sec:budget}. On the backbone axis, IG on GLM-5.3 Flash wins every game against the four Opus~4.6 baselines and both rounds against IG on Opus~4.6 on all three benchmarks, while IG on DeepSeek V4 Flash averages 69\% over SSB and HLE and wins 75\% on APEX-Agents (Appendix~\ref{app:open_models}). IG can therefore be used with open-weight backbones, although its performance depends on the model chosen.

\subsection{Expert Intervention Study}
\label{sec:he_intervention}
The Expert Intervention Study, our primary downstream evaluation, tests whether a report helps practitioners make better scaffold modifications, measured on held-out tasks. We compare IG with CC Subagents, the alternative with the highest pairwise win rate in Section~\ref{sec:llm_judge} and a lightweight architecture practitioners can adopt directly. Participants, all professional software engineers who develop LLM agents, were randomly assigned to a report condition, blinded to its source, and asked to improve the scaffold with standard coding tools within a fixed time budget (1.5 hours on SSB, 2 hours on APEX-Agents); each SSB participant's score is the mean of three evaluation runs over 100 held-out tasks, and each APEX-Agents participant's score, like the unmodified baseline in Table~\ref{tab:he_intervention}, is a single evaluation run over the 144 held-out tasks. Tooling, benchmark access, and time budget are identical across arms, so the report is the only element varied by design. All participants in an arm received the same report. We recruited practitioners rather than crowd workers, prioritizing real-world relevance over sample size (protocol in Appendix~\ref{app:participant_protocol}; statistical analysis in Appendix~\ref{app:stats}).

\begin{table}[t]
\centering
\caption{Expert Intervention Study. Held-out score of the scaffold after $n$ practitioners per arm modified it given one report; the unmodified scaffold is the baseline. Parentheses: standard error of the mean across participants, in percentage points (pp). The SSB human study uses a GPT-5.2 scaffold (Appendix~\ref{app:participant_protocol}). Bold: best per benchmark.}
\label{tab:he_intervention}
\small
\begin{tabular}{llccc}
\toprule
Benchmark & Condition & $n$ & Score & $\Delta$ vs.\ baseline \\
\midrule
\multirow{3}{*}{SSB (pass rate)} & Unmodified scaffold & -- & 27.0\% & -- \\
 & CC Subagents & 6 & 43.2\% ($\pm$4.1) & $+$16.2\,pp \\
 & IG & 6 & \textbf{57.4\%} ($\pm$1.5) & $+$30.4\,pp \\
\midrule
\multirow{3}{*}{APEX-Agents (Pass@1)} & Unmodified scaffold & -- & 19.44\% & -- \\
 & CC Subagents & 8 & 18.40\% ($\pm$0.75) & $-$1.04\,pp \\
 & IG & 8 & \textbf{20.49\%} ($\pm$0.63) & $+$1.05\,pp \\
\bottomrule
\end{tabular}
\end{table}

\paragraph{Results.} On SSB, practitioners given IG reports improved the scaffold by 30.4\,pp over the unmodified baseline, against 16.2\,pp for those given CC Subagents reports (57.4\% vs.\ 43.2\%, Table~\ref{tab:he_intervention}; two-sided Welch's $t$-test $p = 0.016$, Cohen's $d = 1.89$). On APEX-Agents, IG recipients again scored higher, 20.49\% vs.\ 18.40\% Pass@1 ($+2.09$\,pp, Cohen's $d = 1.06$), with the two-sided Welch's $t$-test at $p = 0.053$ for $n = 8$ per arm.

\subsection{Expert Rating Study}
\label{sec:he_judge}
Twelve experts scored 23 insights, every finding of one IG report (11) and one CC Subagents report (12) generated on the SSB training corpus, on a four-dimension 1--5 rubric (correctness, depth, evidence quality, actionability), one insight at a time in a fixed format and blinded to source (Table~\ref{tab:he_judge}, Appendix~\ref{app:he_judge_details}). Aggregate ratings are a near-tie (4.25 for IG vs.\ 4.22 for CC Subagents): IG is ahead on depth ($+0.17$) and evidentiary support ($+0.06$), CC Subagents on correctness ($+0.14$), and actionability is nearly identical. Raters thus score IG higher overall, but none of these differences is statistically significant given the per-insight standard errors in Table~\ref{tab:he_judge}. Agreement between individual raters is low (Krippendorff's $\alpha = 0.10$ across all 12 raters), and the 12-rater mean per insight reported in Table~\ref{tab:he_judge} is only moderately reliable (intraclass correlation coefficient, ICC(2,$k$) $= 0.60$), so the ratings cannot support a sharp claim in either direction. The closest LLM-judge analogue to these ratings is its rubric, and on SSB the rubric also separates the two systems only slightly (composite 1.20 vs.\ 1.12 on the 0--3 scale, Appendix~\ref{app:llm_judge_details}); the judge's large margin is in the pairwise tournament, which compares whole reports rather than individual insights. We therefore treat downstream outcomes, not insight ratings, as the arbiter of report quality: practitioners blinded to report source improved the scaffold by 30.4\,pp with IG reports against 16.2\,pp with CC Subagents reports (Section~\ref{sec:he_intervention}), the same ordering the LLM judge gives.

\subsection{Automated Patcher Loop}
\label{sec:patcher}
We next test whether reports support \emph{iterative} scaffold improvement with no human in the loop. In each round a coding agent (Claude Code with Claude Opus~4.6) modifies the scaffold given the current report and the scaffold source; the patched scaffold collects fresh traces on the training split, the analysis system regenerates its report, and the loop continues until a validation-set patience criterion is met (Appendix~\ref{app:patcher_loop}; patcher prompt in Appendix~\ref{app:patcher_prompt}). We compare IG, RLM, and CC Subagents reports with a pure patcher that receives neither traces nor a report, and also run IG reports from the two open-weight backbones. We run two trials on SSB and one on APEX-Agents, and report each arm's held-out test score at its best validation round, and its best test score as an upper bound (pass rate on SSB; tasks solved of 144 on APEX-Agents).

\paragraph{Results.} On SSB, validation and test track each other across rounds (Pearson $r = 0.91$ and $0.93$ in the two trials), so the validation-selected round is the main result. IG reports are at the top in both trials, tying the pure patcher at 0.80 in the first and leading at 0.879 in the second (Table~\ref{tab:patcher_best}). On APEX-Agents, validation predicts test only weakly ($r = 0.25$), so we report the best test round as an upper bound. IG reports raise the scaffold from 30.8/144 to 42/144, the largest gain of any arm, against 35 for the pure patcher and 32 for CC Subagents and RLM. Validation-selected scores and statistical caveats are in Appendix~\ref{app:patcher_loop}. Beyond the Opus~4.6 arms, we also test whether reports from a cheaper backbone transfer downstream, running one trial per open-weight arm. On SSB, IG on GLM-5.3 Flash matches IG on Opus~4.6, reaching 0.828 at the validation-selected round, between the two Opus~4.6 trials (0.80 and 0.879). On APEX-Agents it reaches 34 of 144 tasks at the best test round: below uncapped IG on Opus~4.6 (42), but comparable to the other Opus~4.6 arms in Table~\ref{tab:patcher_best} (32--35) and equal to IG on Opus~4.6 under a \$25 or \$50 per-round cap (Table~\ref{tab:budget_scaling}), at about \$19 per report. IG on DeepSeek V4 Flash reaches 27, below the unmodified scaffold (30.8), a regression that reflects how hard APEX-Agents is to improve (Appendix~\ref{app:open_models}).

\begin{table}[t]
\centering
\caption{Automated patcher loop: held-out score per arm. SSB: pass rate at the validation-selected round and at the best test round, averaged over two trials. APEX-Agents: best test round (upper bound), tasks solved of 144, uncapped analysis budget. Unmodified APEX-Agents scaffold is pooled over five trials. Pure patcher: no traces or report. Per-trial scores and validation-selected APEX-Agents scores are in Table~\ref{tab:patcher_full}. Bold: best per column.}
\label{tab:patcher_best}
\small
\begin{tabular}{l c c c}
\toprule
Arm & SSB, val-selected & SSB, best test & APEX-Agents, best test \\
\midrule
IG & \textbf{0.84} & \textbf{0.86} & \textbf{42} \\
CC Subagents & 0.79 & 0.82 & 32 \\
RLM & 0.80 & 0.84 & 32 \\
Pure patcher & 0.78 & 0.78 & 35 \\
Unmodified scaffold & 0.41--0.44 & 0.41--0.44 & 30.8 \\
\bottomrule
\end{tabular}
\end{table}

\subsection{Ablations, Cost Scaling, and Backbone Choice}
\label{sec:budget}
\paragraph{Component ablations.} Holding the corpus, backbone, patcher, splits, and a \$50 per-round analysis budget fixed, we removed one IG component at a time: \textbf{Orchestrator-only} disables subagent dispatch, \textbf{Generic Subagents} keeps parallel dispatch but replaces the typed Scout and Investigator with one untyped subagent, \textbf{no Python REPL} removes the agent-writable Python environment and code chaining, so agents reach the data layer only through discrete LLM-mediated tools, and \textbf{no LLM extraction} removes the per-trace extraction tool while the trace table and column persistence remain. Every removal lowers the held-out patcher score below full IG at the same cap (34/144): 32, 31, 30, and 19 tasks respectively. On SSB, every removal also scores below full IG at the validation-selected round (0.828--0.838 against 0.879). Only removing LLM extraction moves the score outside single-trial baseline noise (more than four baseline standard deviations below the unmodified scaffold); the other removals are within it. Table~\ref{tab:ablation_judge} pairs these downstream scores with the judged quality of uncapped reports from the same ablation configurations under both judge families.

\paragraph{Cost scaling.} On the SSB and HLE judge corpora, an IG analysis run averages about \$44 in LLM cost and 48 minutes of wall-clock, against \$8 for RLM, \$26 for Trace2Skill, and \$38 for CC Subagents (Appendix~\ref{app:compute}). To test whether IG's advantage is bought with budget, we ran the APEX-Agents loop with each analysis system's per-round spend capped at \$10, \$25, and \$50, and uncapped, where IG's billed spend on the longer APEX-Agents traces is about \$174 per round. Budget buys breadth and then depth: Scouts per round grow from 1 at \$10 to about 8 from \$50 on, and Investigators first appear at \$50 and reach 12 per round uncapped. Report quality rises with budget under both judge families across the capped tiers (Opus~4.6 composite 1.24, 1.36, and 1.61 at \$10, \$25, and \$50; 1.87 uncapped), while the downstream score stays at 33--34/144 across the capped tiers and reaches 42/144 uncapped (Table~\ref{tab:budget_scaling}). At matched caps of \$10 to \$50, neither IG nor CC Subagents nor RLM separates from the unmodified scaffold (all cells within 27--35/144; Table~\ref{tab:apex_loop}); IG's downstream gain appears when the cap is removed.

\paragraph{Backbone choice.} IG's cost also falls by switching to an open-weight backbone. On GLM-5.3 Flash an IG report costs \$8, \$18, and \$19 on SSB, HLE, and APEX-Agents, against \$17, \$71, and \$174 on Opus~4.6 (Appendix~\ref{app:open_models}); its judged and downstream results are in Sections~\ref{sec:llm_judge} and~\ref{sec:patcher}.

\section{Discussion}
\label{sec:discussion}

\paragraph{Implications for agent development.} The results carry a concrete implication for practitioners developing LLM agents: IG-style corpus-level diagnostic reports support more effective scaffold improvements than competing analysis approaches when consumed by human experts. The 14.2\,pp gap between IG and CC Subagents in the SSB intervention study (Section~\ref{sec:he_intervention}) suggests that report depth and evidentiary grounding matter for practitioner effectiveness, not only topical coverage, a pattern consistent with IG's lead on mechanism and specificity under the LLM judge and with the Investigator role's explicit corpus-scale validation step. By contrast, Single-Agent Coding's reports carry fewer findings and fewer corpus-wide counts, and CC Subagents cross-checks findings only through the orchestrator's summaries, while IG's Investigators re-test each finding against the full corpus. The automated patcher loop (Section~\ref{sec:patcher}) points the same way: at validation-selected checkpoints IG reports match or exceed every other report source on SSB in two independent trials, and on the harder APEX-Agents benchmark they reach the highest best-round score. Report quality also rises with analysis budget (Section~\ref{sec:budget}), so practitioners can trade cost for diagnostic depth.

\paragraph{Limitations and extensions.}
\label{sec:limitations}
Both human studies are small because participants must be experienced agent developers: $n=6$ and $n=8$ per arm in the intervention studies and 23 rated insights in the rating study. Every participant in an arm also received the same report, so each study measures the utility of one report per system rather than variation across report generations. The SSB effect survives leave-one-out and permutation checks and the APEX-Agents effect is large ($d = 1.06$) but does not reach significance ($p = 0.053$); a fully powered replication across more benchmarks is the natural next step. Each APEX-Agents patcher cell is a single trial of a long coding-agent rollout, so differences of a few tasks lie within run-to-run noise, and repeated trials per arm would sharpen the downstream comparisons, including the budget-matched ones. On APEX-Agents, validation also predicts test only weakly, so the patcher results there are best-round upper bounds rather than selected checkpoints. All analysis systems and the primary judge run on Claude Opus~4.6 (Claude Haiku~4.5 for extraction and summarization); the GPT-5.4 judge shows that the ranking in pairwise win rate does not depend on the judge family, and IG keeps its lead on two open-weight backbones (Appendix~\ref{app:open_models}). On scale, no single trace in our corpora exceeds the agent context window; content-aware chunking with per-chunk retrieval decouples per-trace length from context, but preserving cross-chunk dependencies in million-token trajectories remains open. The SSB and HLE corpora hold one trace per task, so per-prompt grouping, which isolates nondeterministic failures across repeated runs of one task, is not exercised here although IG's cohort mechanism supports it. Finally, we evaluate one downstream use of insights, scaffold improvement by practitioners and by an automated patcher. Confirmed failure cohorts and quantified failure patterns are also natural inputs to SFT data filtering, RL reward design, and evaluation-set construction, which we leave to future work.

\section{Conclusion}
We present Insights Generator, a multi-agent system for corpus-level trace diagnostics. We formalize population-level pattern discovery in agent traces as a problem distinct from single-trajectory debugging and taxonomy classification, introduce a scout-investigator architecture with a persisted, schema-cached data layer, and propose a four-setting evaluation framework varying who evaluates and what is measured, applied on three benchmarks under two judge families. IG ranks first on pairwise report quality under both judges and, under the primary judge, leads every rubric dimension. Practitioners given IG reports improve an SSB scaffold by 30.4\,pp, nearly double the gain from the strongest alternative, and an automated patcher given IG reports matches or exceeds every other report source on SSB at validation-selected checkpoints and achieves the largest held-out gain on APEX-Agents, with report quality rising as analysis budget grows. As agent complexity grows, IG provides a systematic framework for analyzing traces, diagnosing failure modes, and generating impactful improvements.

\bibliographystyle{abbrvnat}
\newpage
\bibliography{references}

\newpage
\appendix
\raggedbottom
\section{Appendix}

\subsection{Analysis Loop Pseudocode}
\label{app:algorithm}
\begin{algorithm}[H]
  \caption{Insights Generator analysis loop. Scout and Investigator batches run in parallel; only confirmed findings enter the report.}
  \label{alg:ig_loop}
  \begin{algorithmic}[1]
  \REQUIRE Trace corpus $\mathcal{C} = \{\tau_1, \dots, \tau_n\}$, diagnostic question $Q$
  \STATE $\mathcal{S} \gets \textsc{BuildSchemaCache}(\mathcal{C})$
  \STATE $\mathcal{H}^* \gets \emptyset$ \COMMENT{accumulated candidate hypotheses}
  \STATE $\mathcal{F} \gets \emptyset$ \COMMENT{confirmed findings}
  \WHILE{Orchestrator has not issued \textsc{SubmitReport}}
    \STATE $a \gets \textsc{OrchestratorDecide}(\mathcal{S},\, \mathcal{H}^*,\, \mathcal{F},\, Q)$
    \IF{$a = \textsc{DispatchScouts}$}
      \STATE $\mathcal{H} \gets \textsc{DispatchScoutAgents}(\mathcal{C}_{\mathrm{sample}},\, \mathcal{S},\, Q)$ \COMMENT{parallel; propose}
      \STATE $\mathcal{H}^* \gets \mathcal{H}^* \cup \mathcal{H}$
    \ELSIF{$a = \textsc{DispatchInvestigators}$}
      \STATE $\mathcal{H}_{\mathrm{sel}} \gets \textsc{Prioritize}(\mathcal{H}^*)$
      \STATE $\mathcal{F}_{\mathrm{new}} \gets \textsc{DispatchInvestigatorAgents}(\mathcal{C},\, \mathcal{S},\, \mathcal{H}_{\mathrm{sel}})$ \COMMENT{parallel; validate}
      \STATE $\mathcal{F} \gets \mathcal{F} \cup \{f \in \mathcal{F}_{\mathrm{new}} : f.\mathrm{status} = \texttt{confirmed}\}$
      \STATE $\mathcal{S} \gets \textsc{Consolidate}(\mathcal{S})$ \COMMENT{merge new extractions; refresh schema}
    \ENDIF
  \ENDWHILE
  \STATE $\mathcal{F} \gets \textsc{RemapTraceIDs}(\mathcal{F},\, \mathcal{C})$
  \STATE \textbf{return} $\textsc{SubmitReport}(\mathcal{F},\, Q)$
  \end{algorithmic}
  \end{algorithm}

\subsection{Agent System Prompts}
\label{app:agent_prompts}

The system prompts below correspond to the production configuration in
which both Scout and Investigator subagents are dispatched and the full
IG analysis toolkit is enabled. Ablation variants are rendered
programmatically from the same Jinja templates with conditional blocks
gated on which subagent roles are present.

\paragraph{Orchestrator.} In this prompt, ``scouts'' refers to all dispatched subagents (Scouts and Investigators).

\begin{quote}\small\ttfamily\raggedright
You are the Insights Generator Orchestrator --- an analysis coordinator
that discovers patterns in AI agent execution traces.

\textbf{Your Role.} You find failures (explicit actions, silent errors,
and omissions --- things the agent should have done but didn't),
successes, behavioral differences, efficiency issues, and any noteworthy
patterns across a corpus of agent traces. Your primary job is dispatching
specialized subagents and synthesizing their findings --- you do NOT do
the heavy statistical analysis yourself. You DO have light read-only
tools so you can deep-dive on a search hit when that informs your next
dispatch decision; reserve heavy extraction, cohort comparison, and
cohort-saving for the scouts.

\textbf{Workflow.} (1) Review the schema. (2) Dispatch scout agents
with specific exploration directives. (3) Review hypotheses and
prioritize the most promising ones. (4) Dispatch investigator agents with
hypotheses to validate. (5) Evaluate coverage. (6) Write the final
report.

\textbf{Tools.} A Python REPL for orientation (schema, trace counts,
search, summarization, extraction-column plumbing); plus dispatch tools
for parallel subagent dispatch. Heavy LLM and statistical work is
intentionally not in the orchestrator's toolkit.

\textbf{Dispatching subagents.} Directives focus on specific areas
(``verbose database queries correlate with null results'', not ``look at
tool calls''). Mix directive shapes across each batch: some target active
errors (wrong action, bad output, failure text), others target expected
actions or checks the agent may have OMITTED. Dispatches return compact
summaries with full per-hypothesis evidence written to a sandbox sidecar
that the orchestrator can read via the REPL.

\textbf{Anti-overfitting.} Do not generalize from a single scout
agent's sample. Every finding must be validated by an investigator agent with
statistical evidence. Small-cohort findings are often high-value; do not
drop a hypothesis because its cohort is small.

\textbf{Iteration.} The pipeline is flexible: each dispatch tool may be
called multiple times across the run, in any order. Default toward
iterating; submit only when no credible lead remains undispatched.
\end{quote}

\paragraph{Scout (Hypothesis Agent).}

\begin{quote}\small\ttfamily\raggedright
You are a Hypothesis Agent (Scout) --- you explore traces of agent
execution to discover patterns and generate hypotheses.

\textbf{Your role.} Look at trace data to find interesting patterns the
orchestrator should investigate further. Emphasize breadth over depth ---
scan many traces to identify potential issues, not deeply validate any
single one.

\textbf{What to look for.} Explicit failures, silent failures (agent
completes but does a poor job), efficiency issues, success patterns,
behavioral differences, and OMISSIONS (steps the agent should have taken
but didn't). Do not limit yourself to a failure taxonomy --- discovery is
the goal.

\textbf{Sampling guidance.} Use representative samples --- typically
50--200 traces --- to form hypotheses.

\textbf{Workflow.} (1) Orient via schema and a small Polars pass to
understand data shape. (2) Discover via search, issuing per-concept
queries (failures, successes, loops, escalations, possible omissions).
(3) Read samples on interesting hits. (4) Quantify by extracting on small
focused samples (10--30 traces) to test patterns. Before extracting,
check what's already been extracted by other agents. (5) Submit
hypotheses.

\textbf{Searching for omissions.} Search cannot match absence directly.
Imagine the text an agent that HAD performed the expected action would
produce, search for THAT, then contrast cohorts in Polars or in a
follow-up search.

\textbf{Output.} A JSON list of hypotheses, each with name, description,
evidence (trace id, verbatim quote, and explanation), and estimated
prevalence. Optionally include 1--3 well-motivated suggestions for
further investigation that this Scout did not have time or scope to
pursue.
\end{quote}

\paragraph{Investigator (Explore Agent).}

\begin{quote}\small\ttfamily\raggedright
You are an Explore Agent (Investigator) --- you validate hypotheses
about agent execution traces with statistical evidence.

\textbf{Your role.} Take a specific hypothesis and validate it with
rigorous evidence from the full trace corpus (or a large representative
subset). Emphasize depth over breadth --- quantify patterns, run
comparisons, and provide statistical backing.

\textbf{Validation approach (hierarchical).}
Level A --- broad stats: compute statistics across all traces using
Polars group-bys.
Level B --- focused comparison: extract relevant features on 50--100
traces; check existing extractions first to avoid duplicating work.
Level C --- deep comparison: run cohort comparisons 2--3 times with
different samples for robust findings.

\textbf{Output.} A single submission with status (confirmed, refuted, or
inconclusive), 1--2 sentence summary, quantified prevalence, and 8--10+
evidence items. Each evidence item contains a trace id, a verbatim quote
from the trace, and a 2--3 sentence explanation that names the specific
mechanism visible in THIS trace and connects it to the broader pattern.
Hypothesis-test outputs (p-values, Cram\'er's V, odds ratios) go in
additional observations, not the main summary. Provide a single concrete
suggested action or null.

\textbf{Anti-overfitting.} Do not generalize from fewer than 20 traces in
a comparison. Quantify everything. Flag confidence level: high (more than
100 traces), medium (20--100), low (fewer than 20).

\textbf{Saving the affected cohort.} Record the defensible cohort --- the
subset you are confident exhibits the pattern, NOT the broader pool you
analyzed. Submission is rejected if this has not been called.
\end{quote}

\subsection{Implementation Details}
\label{app:implementation_details}

Table~\ref{tab:ig_config} lists the production IG configuration used in
all reported experiments. All fields correspond directly to attributes of
the IG configuration dataclass; ablation experiments override the
relevant fields and leave the rest at their defaults.

\begin{table}[H]
\centering
\small
\caption{IG production configuration. Concurrency, timeout, and sandbox
parameters are those used for the runs reported here and may be lowered for
cost-constrained deployments.}
\label{tab:ig_config}
\begin{tabular}{lll}
\toprule
Group & Field & Value \\
\midrule
Models & Orchestrator           & Claude Opus 4.6 \\
       & Hypothesis (Scout)     & Claude Opus 4.6 \\
       & Explore (Investigator) & Claude Opus 4.6 \\
       & Cohort comparison      & Claude Opus 4.6 \\
       & Per-trace extraction   & Claude Haiku 4.5 \\
       & Trace summarization    & Claude Haiku 4.5 \\
\midrule
Extraction & Concurrency             & 50 parallel calls \\
           & Checkpoint interval     & 5{,}000 traces \\
           & Auto-save threshold     & 250 traces \\
           & Per-call timeout        & 300\,s \\
\midrule
Orchestration & Max orchestrator turns & 500 (self-terminates via report) \\
              & Max subagent turns     & 500 (self-terminate via submit) \\
\midrule
Traces & Long-trace chunk threshold & 50{,}000 tokens \\
       & Retrieval top-$k$           & 20 (search default) \\
\midrule
Ablation & Mode             & full \\
         & Tool backend     & REPL \\
         & REPL state       & stateful \\
\midrule
Sandbox & CPU       & 16 cores \\
        & Memory    & 65{,}536\,MB (64\,GB) \\
        & Timeout   & 14{,}400\,s (4\,h) \\
\bottomrule
\end{tabular}
\end{table}

All LLM calls route through a centralized LLM proxy. Subagents run in a shared sandboxed container with the analysis primitives pre-injected into a stateful Python REPL session; the orchestrator and each subagent get their own session.

\subsection{Insights Generator Tools}
\label{app:tools}

Table~\ref{tab:tools} lists the 14 IG analysis primitives organized by
category, with per-role availability across the Orchestrator, Scout
(Hypothesis Agent), and Investigator (Explore Agent).

\begin{table}[H]
\centering
\small
\caption{IG analysis primitives, pre-injected as Python functions in every agent's REPL session and called from agent-written code. \cmark{}/\xmark{}: available or not to the Orchestrator (Orch.), Scout, and Investigator (Inv.).}
\label{tab:tools}
\begin{tabular}{l>{\raggedright\arraybackslash}p{6.0cm}ccc}
\toprule
Function & Description & Orch. & Scout & Inv. \\
\midrule
\multicolumn{5}{l}{\emph{Trace inspection}} \\
\texttt{load\_traces} & Load full corpus + extractions as a Polars DataFrame & \cmark & \cmark & \cmark \\
\texttt{get\_trace} & Return full content of one trace; chunk-list outline if oversized & \cmark & \cmark & \cmark \\
\texttt{get\_trace\_chunk} & Retrieve a specific chunk of a long trace & \cmark & \cmark & \cmark \\
\texttt{get\_schema} & Per-column metadata: types, null rates, distributions, correlations & \cmark & \cmark & \cmark \\
\texttt{search\_traces} & Hybrid semantic-keyword retrieval over trace content & \cmark & \cmark & \cmark \\
\texttt{get\_extractions} & List previously persisted extraction columns and their field definitions & \cmark & \cmark & \cmark \\
\midrule
\multicolumn{5}{l}{\emph{LLM-driven analysis}} \\
\texttt{extract} & Structured feature extraction across traces with chunking, batching, and checkpointing & \xmark & \cmark & \cmark \\
\texttt{summarize\_trace} & Per-trace LLM summary with caching and progressive summarization for long traces & \cmark & \cmark & \cmark \\
\texttt{summarize\_group} & Single LLM summary characterizing a group of traces & \cmark & \cmark & \cmark \\
\texttt{compare\_segments} & Deep A/B comparison between two cohorts with token budgeting & \xmark & \xmark & \cmark \\
\midrule
\multicolumn{5}{l}{\emph{Data management}} \\
\texttt{save\_column} & Persist a computed Polars column as an extraction parquet & \cmark & \cmark & \cmark \\
\texttt{save\_affected\_traces} & Persist a trace-id list as the cohort supporting a finding & \xmark & \xmark & \cmark \\
\texttt{reload\_data} & Refresh cached DataFrame after new extractions land & \cmark & \cmark & \cmark \\
\texttt{consolidate} & Merge extraction parquets into the base store and refresh the schema cache & \cmark & \cmark & \cmark \\
\bottomrule
\end{tabular}
\end{table}

\subsection{Benchmark and Corpus Statistics}
\label{app:benchmark_details}

The experiments in this paper use three benchmarks:
\textbf{SpreadsheetBench (SSB)}~\citep{spreadsheetbench},
\textbf{HLE}~\citep{hle}, and \textbf{APEX-Agents}~\citep{apexagents}. For each benchmark, traces were collected by
executing a fixed ReAct-style scaffold over the benchmark's task set;
all downstream analysis, ablation, and intervention experiments reuse
those fixed corpora rather than re-collecting traces per system. The
SSB human evaluation experiments
(Sections~\ref{sec:he_intervention} and~\ref{sec:he_judge}) use traces
generated by \texttt{openai/gpt-5.2-2025-12-11}, while the LLM judge
(Section~\ref{sec:llm_judge}) and patcher-loop
(Section~\ref{sec:patcher}) experiments, and all APEX-Agents experiments, use traces generated by
\texttt{openai/gpt-5.4-2026-03-05}. The sole exception is the
automated patcher loop, which, by construction, regenerates a fresh
trace corpus each round from the patched scaffold $\mathcal{A}_r$.
Corpus-level statistics for all three corpora are summarized in
Table~\ref{tab:corpus_stats}.

\paragraph{SpreadsheetBench (SSB).}
SpreadsheetBench tasks require an agent to read an Excel workbook,
perform a manipulation specified in natural language (e.g.\ filter,
summarize, transform, write a formula), and persist the result back to
the workbook. Outputs are graded by exact cell-value match against
ground truth, which makes the benchmark deterministically scorable. SSB is the primary benchmark for the human studies because its tasks
are familiar to technology professionals, run quickly, and admit
prompt, tool, and control-flow fixes. The
scaffold is a ReAct agent with a 5-turn budget that exposes
\texttt{openpyxl} and \texttt{pandas} via a Python code-execution tool;
the agent has no web access. The LLM judge corpus
(\texttt{ssb\_gpt-5.4-med\_20260430}, used in
Section~\ref{sec:llm_judge}) contains 296 traces with a baseline pass
rate of 42.2\% (125 correct, 171 incorrect). Tasks are categorized by
the benchmark into \textit{Cell-Level} and \textit{Sheet-Level}
manipulation, with markedly different difficulty profiles
(approximately 27\% vs.\ 76\% pass rate in our corpus). The
human-expert experiments (Sections~\ref{sec:he_judge}
and~\ref{sec:he_intervention}) use the SSB training-split corpus of 250
traces (110 correct, 140 incorrect; baseline pass rate 44\%); the
training-split bundle is described in detail in
Appendix~\ref{app:exp1_protocol}. The automated patcher loop
(Section~\ref{sec:patcher}) collects fresh training-split traces on
every round from the patched scaffold; the round-0 baseline pass rate
is 0.41 on the held-out test split.

\paragraph{HLE.}
HLE (Humanity's Last Exam) is an expert-level question-answering benchmark spanning physics, mathematics, biology,
chemistry, and humanities, where each task requires the agent to return
a single short answer along with a self-reported confidence in
$[0, 100]$. The scaffold is a ReAct agent that exposes web search,
web-page fetch, and Python code execution; turn counts in our corpus
range from 0 (instant-answer / hang) to 18 with a long-tail
distribution. The LLM judge corpus
(\texttt{hle\_gpt-5.4\_20260430}) contains 250 traces with a
baseline pass rate of 23.6\% (59 correct, 191 incorrect); 207 of 250
traces use at least one tool. The 43 zero-tool traces achieve only
9.3\% accuracy versus 26.6\% for tool-using traces, a sub-population
gap that surfaces consistently across systems' findings.
Correctness for HLE is graded by an LLM judge applied to the
benchmark's reference answers (rather than by deterministic match), so
HLE serves as a complementary stressor to SSB's deterministic grader.

\paragraph{APEX-Agents.}
APEX-Agents contains long-horizon, document-grounded professional
tasks in law, investment banking, and management consulting. Relative
to SSB its traces contain roughly 13$\times$ more tool calls, its tasks
expose over 100 more files and over 10 more tools each, and each trial
produces about 4$\times$ more output tokens. Failures we observed include
long-horizon planning errors, tool-output context exhaustion, and silent
data-processing errors, alongside isolated API misuse. The
Expert Intervention Study used a 236-trace corpus, a 20-task development
split for participant iteration, and a 144-task held-out test split; the
automated patcher loop and budget grid report strict Pass@1 on the same
144 held-out tasks against an unmodified-scaffold baseline pooled over
five trials (154/720, 0.214). The scaffold is a ReAct-style agent (GPT-5.4, medium reasoning effort) with filesystem search and read, spreadsheet, document, slide, and PDF reader tools plus code execution and no web access, a 100-turn budget, and a one-hour wall-clock limit; 215 of the 236 traces call at least one tool, and 20 of the 21 zero-tool traces are law tasks.

\begin{table}[H]
\centering
\caption{Trace corpora for SSB, HLE, and APEX-Agents. Eval-corpus traces come from a generic ReAct scaffold on \texttt{openai/gpt-5.4-2026-03-05}; the SSB human-study split uses GPT-5.2 traces (Appendix~\ref{app:participant_protocol}). Eval corpus: the corpus released to every analysis system for the LLM-judge evaluation (Section~\ref{sec:llm_judge}); human-study and patcher splits are listed separately. Parenthesized trace counts are correct / incorrect.}
\label{tab:corpus_stats}
\scriptsize\setlength{\tabcolsep}{3pt}
\begin{tabular}{@{}p{3.1cm} p{3.35cm} p{3.35cm} p{3.35cm}@{}}
\toprule
                              & \textbf{SSB}            & \textbf{HLE} & \textbf{APEX-Agents} \\
\midrule
Citation                      & \citet{spreadsheetbench}             & \citet{hle} & \citet{apexagents} \\
Domain                        & Spreadsheet manipulation             & Expert QA with web search & Long-horizon professional tasks \\
Eval metric                   & Exact cell-value match               & LLM-judged correctness & Pass@1 (rubric grader) \\
Eval-corpus traces            & 296 (125 correct / 171 incorrect)    & 250 (59 correct / 191 incorrect) & 236 (47 correct / 189 incorrect) \\
Eval-corpus pass rate         & 42.2\%                               & 23.6\% & 19.9\% \\
Tool-using traces             & all (code-exec mandatory)            & 207 / 250 (82.8\%) & 215 / 236 (91.1\%) \\
Tool surface                  & \texttt{openpyxl}, \texttt{pandas}    & web search, web fetch, code-exec & filesystem, spreadsheet, docs, slides, PDF, code-exec \\
Turn budget                   & 5                                    & 20 (observed max 18) & 100 (1\,h wall-clock) \\
\midrule
Human-study split             & 250 train (110 / 140), 44\% pass     & ---  & 236 train, 20 dev, 144 test \\
\midrule
Replicates per system & 3 independent re-runs       & 3 independent re-runs & 1 \\
Union-gold clusters           & 50                                   & 50 & 35 (Opus~4.6), 27 (GPT-5.4) \\
\bottomrule
\end{tabular}
\end{table}

\paragraph{Replicates, gold clusters, and corpus size.}
For the LLM judge evaluation each system is run 3 times on SSB and HLE (once on APEX-Agents)
on the same fixed corpus to capture run-to-run variance, and
the corresponding union-gold reference set per benchmark is built by
pooling findings across all systems and replicates and clustering them with
the meta-evaluator prompt in Appendix~\ref{app:llm_judge_prompt};
clustering produced 50 canonical observations on SSB and HLE and 35 on APEX-Agents (27 under the GPT-5.4 judge) (the clustering imposes no cluster cap),
which is the cluster count against which coverage and per-dimension
quality are scored in Tables~\ref{tab:judge_by_benchmark} and~\ref{tab:llm_judge_combined}. No single trace exceeds the agent context window, but the
full 296-trace SSB corpus and 250-trace HLE corpus each exceed any
single LLM context, motivating the corpus-level analysis primitives
described in Section~\ref{sec:ig}.

\subsection{Comparison System Implementation Details}
\label{app:baseline_details}

All five systems answer the same canonical diagnostic question
(Appendix~\ref{app:llm_judge_prompt}) over the same trace corpus, and all
emit a session artifact normalized to the same JSON schema by a shared
report-utils module. The systems differ along three axes:
\textit{tool surface} (raw filesystem and coding tools vs.\ IG's
structured analysis primitives), \textit{topology} (single agent vs.\
multi-agent dispatch), and \textit{data layer} (raw text vs.\ structured
Polars store with extraction columns). All systems use Claude Opus~4.6
as their backbone agent model.

\paragraph{Single-Agent Coding.} A single Claude Code CLI session,
invoked headless, with a filesystem-based knowledge layout (one file per
trace plus an index) and Claude Code's native tool surface (read, write,
edit, glob, grep, bash). The agent discovers patterns through filesystem
retrieval, ad-hoc scripting, and manual aggregation. It establishes the floor for what a strong coding agent can achieve without a dedicated
trace-analysis scaffold.

\paragraph{CC Subagents.} Claude Code with subagent dispatch via the
\texttt{Task} tool, implementing the same scout-investigator role
decomposition as IG. Each subagent receives a focused directive from the
orchestrator and returns a compact summary; subagents do not share a
structured data layer and each runs in an isolated context, so all
cross-agent integration happens via the orchestrator's reasoning.

\paragraph{Trace2Skill-style.} A pipeline modeled after
Trace2Skill~\citep{trace2skill}: a flat parallel fleet of analyst
subagents over individual trajectories, followed by a hierarchical
consolidation pass that merges per-batch findings into a conflict-free
output.

\paragraph{RLM.} The vanilla Recursive Language Model implementation
from~\citet{rlm}, wrapped to load a parquet of agent traces and produce a
structured-report JSON. The root LM dispatches recursive sub-LM calls at
maximum depth 2; sub-LMs at maximum depth fall back to plain LM calls.
Distinct from HALO~\citep{halo}, which extends an RLM with a closed-loop
coding-agent stage (Section~\ref{sec:related_work}).

\paragraph{IG (this work).} The full production IG configuration
described in Section~\ref{sec:ig} and
Appendix~\ref{app:implementation_details}, with the orchestrator,
scout, and investigator agents operating over a shared structured trace
store with schema-aware extraction primitives (Algorithm~\ref{alg:ig_loop}).

\subsection{LLM Judge: Evaluation Prompts and Rubrics}
\label{app:llm_judge_prompt}

The LLM judge evaluation in Section~\ref{sec:llm_judge} is layered
into three stages, each using a separate judge prompt. All three stages
use the same judge model (Claude Opus 4.6). The shared
canonical diagnostic question (the input given to every system) is also
reproduced below.

\paragraph{Canonical diagnostic question.} Every comparison system received
the following question over the same trace corpus:

\begin{quote}\small\itshape\raggedright
This corpus contains execution traces from an AI agent attempting the
\{benchmark name\} benchmark. Surface the notable behavioural patterns in the
corpus: recurring errors, silent failures, systematic efficiency issues,
unexpected strengths, and how sub-populations of traces differ (e.g.\
correct vs.\ incorrect, high- vs.\ low-confidence, tool-using vs.\ not).
For every finding: quantify prevalence over the total number of traces in
the corpus; cite at least 3--5 specific trace IDs and direct quotes as
evidence; compare relevant sub-populations; and distinguish confirmed /
refuted / inconclusive. Do not focus on headline benchmark metrics
(accuracy, pass-rate); focus on agent behaviours. Return the structured
JSON report as specified.
\end{quote}

\paragraph{Stage 1: Union-gold clustering.} A single judge call
ingests every finding from every system and clusters them by underlying
observation. The output is a canonical reference set against which each
system is later scored. Prompt:

\begin{quote}\small\ttfamily\raggedright
You are a meta-evaluator clustering findings from multiple automated
trace-analysis systems that all analyzed the SAME corpus of AI-agent
execution traces. Each input finding is a claim about the corpus. Your
job is to produce a `union gold' set: the canonical list of distinct
observations about the corpus that appear across ANY of the systems.

Rules: (1) Two findings belong in the same cluster iff they describe the
same underlying phenomenon in the corpus (same behavior, failure mode,
or pattern), even if their framing, prevalence numbers, or prescriptions
differ. Use the affected trace-id lists as strong evidence for sameness.
(2) A cluster may contain a single finding (unique observations are
valuable). (3) Give each cluster a crisp canonical name (at most 80
characters) and a 1--3 sentence description. (4) Include up to 5
representative trace ids that most strongly evidence the cluster (prefer
ids cited by multiple systems). (5) Output VALID JSON. No prose before
or after the JSON block.
\end{quote}

\paragraph{Stage 2: Per-cluster rubric scoring.} For each system, a
separate judge call scores the system's report against each union-gold
cluster on five dimensions. Each dimension is scored on a 0--3 scale with
explicit anchor descriptions:

\begin{quote}
\small
\begin{tabular}{l>{\raggedright\arraybackslash}p{0.62\linewidth}}
\textbf{detection}     & 0 absent; 1 passing mention; 2 explicit finding; 3 explicit finding given prominence \\
\textbf{mechanism}     & 0 none; 1 vague; 2 plausible; 3 sharp and testable \\
\textbf{evidence}      & 0 none; 1 single; 2 multiple; 3 quantified across corpus \\
\textbf{specificity}   & 0 generic; 1 mildly specific; 2 specific; 3 precise with quantified scope \\
\textbf{actionability} & 0 none; 1 vague; 2 concrete; 3 concrete and testable \\
\end{tabular}
\end{quote}

The judging-discipline instructions explicitly direct the judge to:
score 0 when a dimension is not evidenced in the report (do not give
credit for what a well-intentioned reader might infer); cap detection at
1 if the report makes a factually wrong claim about the observation; and
use the report's own findings verbatim rather than rewarding paraphrase.
The main-body quality table (Table~\ref{tab:llm_judge_combined}) reports
mean scores on the three most discriminative dimensions
(\textit{mechanism}, \textit{specificity}, \textit{actionability}) plus a
composite; \textit{detection} is omitted from the tables, which report coverage instead (Appendix~\ref{app:llm_judge_details}); \textit{evidence} is omitted from
the tables because it is highly correlated with mechanism in our runs.

\paragraph{Stage 3: Pairwise ranking.} For each unordered pair of
systems, the judge is presented with both reports twice in opposite
positions to control for position bias~\citep{mtbench}. Each round is
scored independently. The pairwise judge prompt:

\begin{quote}\small\ttfamily\raggedright
You are an expert ML research engineer comparing two automated
trace-analysis reports that were produced by different systems analyzing
the SAME corpus of AI-agent execution traces and answering the SAME
canonical question.

Your job: pick the report a scaffold-improvement engineer would find
more useful for actually fixing the agent. Weigh specificity, evidentiary
grounding (cited trace ids and quantitative prevalence), novelty
(surfacing non-obvious patterns), and actionability (concrete scaffold
changes). Penalize generic platitudes, unsupported claims, and reports
that confuse breadth for depth.

Answer in strict JSON, with fields: winner (A, B, or tie); confidence
(low, med, or high); rationale (1--3 sentences).
\end{quote}

The win rate reported in Table~\ref{tab:llm_judge_combined} is computed
per system as $(\text{wins} + 0.5 \cdot \text{ties}) / n_{\text{rounds}}$
across all rounds in which the system participated.

\subsection{LLM Judge: Per-Benchmark Breakdowns, Ablations, and Cross-Family Judge}
\label{app:llm_judge_details}

This appendix provides per-benchmark breakdowns of the LLM judge
results summarized in Table~\ref{tab:llm_judge_combined}. All SSB and HLE metrics
are means across three independent report replicates per system; APEX-Agents rows use one report per system.
The four comparison systems are described in Section~\ref{sec:llm_judge}, along with the IG architecture ablations.

\paragraph{Coverage.}
Table~\ref{tab:judge_by_benchmark} reports the coverage metric omitted from
Table~\ref{tab:llm_judge_combined}: the percentage of union-gold
clusters (50 on SSB and HLE, 35 on APEX-Agents) for which the report cites at least one of the traces cited by the cluster's findings.

\begin{table}[H]
\centering
\caption{LLM-judge results by benchmark (Claude Opus~4.6 judge). \textbf{Cov.}: coverage (\%), the share of union-gold clusters for which the report cites at least one trace that the cluster's findings cite (50 clusters on SSB and HLE, 35 on APEX-Agents). \textbf{WR}: pairwise win rate (\%). Mech., Spec., Act., Comp.: 0--3 rubric scores for mechanism, specificity, actionability, and their mean. SSB and HLE: mean of three replicates; APEX-Agents: one report per system.}
\label{tab:judge_by_benchmark}
\small
\begin{tabular}{ll cc cccc}
\toprule
System & Bench. & Cov. & WR & Mech. & Spec. & Act. & Comp. \\
\midrule
\multirow{3}{*}{IG} & SSB & 87 & 70.3 & 1.23 & 1.23 & 1.13 & 1.20 \\
 & HLE & 95 & 81.8 & 1.45 & 1.44 & 1.41 & 1.43 \\
 & APEX & 100 & 88.9 & 1.77 & 1.60 & 1.43 & 1.60 \\
\midrule
\multirow{3}{*}{CC Subagents} & SSB & 93 & 61.5 & 1.15 & 1.15 & 1.05 & 1.12 \\
 & HLE & 97 & 59.9 & 1.10 & 1.07 & 1.09 & 1.09 \\
 & APEX & 97 & 77.8 & 1.26 & 1.20 & 1.23 & 1.23 \\
\midrule
\multirow{3}{*}{Trace2Skill} & SSB & 96 & 54.7 & 1.06 & 1.03 & 1.08 & 1.06 \\
 & HLE & 98 & 47.9 & 1.30 & 1.25 & 1.38 & 1.31 \\
 & APEX & 97 & 61.1 & 0.97 & 0.97 & 0.89 & 0.94 \\
\midrule
\multirow{3}{*}{RLM} & SSB & 89 & 12.5 & 0.95 & 0.95 & 0.81 & 0.90 \\
 & HLE & 96 & 8.9 & 0.96 & 0.94 & 0.87 & 0.92 \\
 & APEX & 100 & 22.2 & 1.26 & 1.29 & 1.03 & 1.19 \\
\midrule
\multirow{3}{*}{Single-Agent Coding} & SSB & 65 & 16.7 & 0.83 & 0.79 & 0.65 & 0.76 \\
 & HLE & 95 & 12.5 & 0.78 & 0.78 & 0.80 & 0.79 \\
 & APEX & 97 & 0.0 & 0.40 & 0.46 & 0.49 & 0.45 \\
\midrule
\multicolumn{8}{l}{\textit{IG architecture ablations}} \\
\multirow{3}{*}{IG, Generic Subagents} & SSB & 87 & 57.3 & 1.02 & 1.02 & 0.89 & 0.98 \\
 & HLE & 93 & 74.0 & 1.26 & 1.23 & 1.17 & 1.22 \\
 & APEX & 97 & 72.2 & 1.57 & 1.40 & 1.31 & 1.43 \\
\midrule
\multirow{3}{*}{IG, Orchestrator-only} & SSB & 78 & 31.8 & 0.92 & 0.90 & 0.79 & 0.87 \\
 & HLE & 86 & 28.1 & 0.81 & 0.83 & 0.85 & 0.83 \\
 & APEX & 77 & 38.9 & 0.83 & 0.80 & 0.69 & 0.77 \\
\bottomrule
\end{tabular}
\end{table}

\paragraph{Pairwise win-rate rankings.}
Table~\ref{tab:judge_by_benchmark} reports per-benchmark pairwise win rates. IG leads on all three benchmarks, and its lead over the next system is larger on HLE (81.8\% vs.\ 59.9\%) than on SSB (70.3\% vs.\ 61.5\%). Each pair is judged in two position-swapped rounds and each round is tallied independently: a pair whose two verdicts disagree contributes one win and one loss to each system, and a verdict the parser cannot read is recorded as a tie.

\paragraph{Per-benchmark rubric scores.}
Table~\ref{tab:judge_by_benchmark} breaks out the per-dimension
quality scores by benchmark. The cross-benchmark averages reported in
Table~\ref{tab:llm_judge_combined} are the means of these rows.

IG's quality advantage is larger on HLE (composite 1.43 vs.\ next-best
Trace2Skill at 1.31) than on SSB (1.20 vs.\ CC Subagents at 1.12),
consistent with the wider pairwise gap on HLE. Trace2Skill's
actionability score is notably strong on HLE (1.38), approaching IG
(1.41); on HLE this is the only dimension where another system comes within 0.03 of IG.

\paragraph{IG architecture ablations: per-benchmark detail.}
Table~\ref{tab:judge_by_benchmark} reports the per-benchmark breakdown of the
two IG architecture ablations summarized in
Table~\ref{tab:llm_judge_combined}. Both ablations preserve IG's full analysis toolkit (Appendix~\ref{app:tools}; the Orchestrator-only variant gives the Orchestrator the Investigator's tool set) and stateful Python data layer;
they differ only in subagent topology. The \textbf{Orchestrator-only}
variant disables subagent dispatch entirely, forcing the orchestrator to
run all extraction, summarization, and cohort comparison itself within its
own context window. The \textbf{Generic Subagents} variant retains parallel
subagent dispatch but replaces the typed Scout (breadth-first hypothesis
proposal) and Investigator (depth-first corpus-scale validation) roles with
a single generic analyst-subagent role that receives an unstructured
analysis directive.

The Orchestrator-only variant degrades sharply on every benchmark
(SSB and HLE means: composite $1.32 \to 0.85$, pairwise WR $76.0\% \to 29.9\%$), and its
coverage also drops below the multi-agent floor reached by every other
multi-agent system in Table~\ref{tab:judge_by_benchmark} (82\% vs.\
$\geq 90\%$). Without parallel subagents, the orchestrator cannot afford
to perform repeated corpus-scale extractions inside its own context, so
findings consolidate around a smaller subset of the union-gold clusters.

The Generic Subagents variant is intermediate: coverage and HLE pairwise
WR remain high (90\% Avg coverage, 74\% HLE WR), but per-dimension
quality drops by roughly 0.20 in composite. The pattern is consistent
across mechanism, specificity, and actionability rather than localized to
any one dimension, suggesting that the typed Scout/Investigator
decomposition affects \emph{how grounded} a finding is (depth of
mechanism, prevalence quantification, citation specificity) rather than
\emph{whether} the finding is surfaced at all. Together, the two
ablations order the configurations as Orchestrator-only $<$
Generic Subagents $<$ full IG, with pairwise WR differences of approximately $35.7$\,pp (Orchestrator-only $\to$
Generic Subagents) and $10.4$\,pp (Generic Subagents $\to$ full IG);
because win rates depend on the tournament pool, these differences are not an additive decomposition of IG's $76.0\%$.

\paragraph{Cross-family judge.}
To test whether the ranking depends on the judge sharing a model family
with the systems under evaluation, we re-ran all three judge stages
(union-gold construction, per-cluster rubric scoring, and the pairwise
tournament) with GPT-5.4 as the judge on the same reports and
the same rubric and prompts. The 10 judged configurations are the seven in Table~\ref{tab:llm_judge_combined} plus three IG tool ablations (no LLM extraction, no Python REPL, no vector search); these three additional configurations enter only the rank correlation and are not tabulated. Table~\ref{tab:cross_judge} reports per-system win rates under the GPT-5.4 judge; on APEX-Agents, ten configurations were judged with one report each (the seven tabulated plus the three IG tool ablations).

\begin{table}[H]
\centering
\caption{Cross-family judge: GPT-5.4 re-judges the reports of Table~\ref{tab:llm_judge_combined} with the same rubric and prompts. \textbf{WR}: pairwise win rate (\%); Avg: mean over SSB and HLE. \textbf{Comp.}: rubric composite (0--3) averaged over SSB and HLE, or on APEX-Agents alone. APEX-Agents: one report per system. Bold: best per column. Last row: Spearman rank correlation between the GPT-5.4 and Opus~4.6 rankings of all judged configurations (10 on each benchmark).}
\label{tab:cross_judge}
\small
\begin{tabular}{l cccc cc}
\toprule
& \multicolumn{4}{c}{Pairwise WR (\%)} & \multicolumn{2}{c}{Comp.\ (0--3)} \\
\cmidrule(lr){2-5} \cmidrule(lr){6-7}
System & SSB & HLE & Avg & APEX & SSB+HLE & APEX \\
\midrule
IG                    & \textbf{77.1} & \textbf{82.8} & \textbf{79.9} & \textbf{94.4} & 1.17 & 1.48 \\
CC Subagents          & 59.4 & 69.3 & 64.3 & 88.9 & 1.07 & 1.38 \\
Trace2Skill           & 43.8 & 50.0 & 46.9 & 50.0 & \textbf{1.25} & 1.37 \\
RLM                   & 16.7 & 10.9 & 13.8 & 44.4 & 0.98 & 1.27 \\
Single-Agent Coding   & 35.4 & 14.6 & 25.0 & 11.1 & 0.86 & 0.63 \\
\midrule
\multicolumn{7}{l}{\textit{IG architecture ablations}} \\
IG, Generic Subagents & 52.6 & 78.1 & 65.4 & 72.2 & 1.16 & \textbf{1.69} \\
IG, Orchestrator-only & 33.9 & 30.7 & 32.3 & 38.9 & 1.00 & 1.00 \\
\midrule
Rank agreement with Opus~4.6 ($\rho$) & 0.99 & 0.95 & -- & 0.77 & \multicolumn{2}{c}{} \\
\bottomrule
\end{tabular}
\end{table}

\paragraph{Faithfulness audits.} Citation validity is checked by resolving every cited trace ID against the corpus and every quoted snippet against the cited trace: 951/951 cited traces resolve on SSB and HLE, while on APEX-Agents citation validity falls to 81.6\% at the lowest analysis-budget cap and reaches 99.8\% uncapped (Appendix~\ref{app:patcher_loop}); verbatim quote match is 62.0\% on SSB and 41.2\% on HLE, and the remaining quotes are not verbatim matches, which we treat as paraphrases of the cited trace but cannot confirm by this check. Prevalence claims are recomputed from the corpus's recorded execution labels (pass/fail, task type, tool-call and runtime counts) for every checkable figure in each report; every one matched exactly or, for continuous rates, to three decimal places, and for the 84\% finding in Section~\ref{sec:intro_example} the denominator (incorrect Python-using traces) reproduces exactly from the labels, while the numerator, traces exhibiting a silent computation failure, is a semantic judgment this check cannot verify. Whether a cited trace exhibits the described mechanism is not checked deterministically by this audit; an LLM entailment check over cited traces is reported in Appendix~\ref{app:filter} but not treated as ground truth. On reference independence, the pairwise tournament uses no gold, and IG contributed findings to only 12 of the 50 SSB union-gold clusters, so 76\% of the reference was derived from other systems' findings; this bounds IG-specific circularity but not circularity in general, since the reference is built from the evaluated systems' outputs, and coverage credits a report whenever it cites one of a cluster's traces.

\subsection{Statistical Baselines}
\label{app:stat_baselines}
We built four baselines from the same traces IG reads. (1) Liblit statistical fault localization~\citep{liblit2005} ranks trace elements (tool calls, error messages, text $n$-grams) by how much more often they occur in failing than passing traces. (2) Combined statistics runs significance tests with multiple-testing correction on 40--60 hand-derived trace features per benchmark, finds subgroups where failures concentrate, and clusters similar traces. (3) Retrieved traces + LLM write-up selects 15 failing and 5 passing traces by BM25 on the failure terms from (1) and writes a report from them in one Claude Opus~4.6 call. (4) Statistics + LLM write-up writes up the outputs of (1) and (2) in one Opus~4.6 call. Each report uses IG's format, cites real trace IDs, and is judged pairwise against every fold-1 report in the pool (the first of the three report replicates per configuration on SSB and HLE; APEX-Agents has one report per configuration) under both judges with the same prompts and citation audit (Table~\ref{tab:stat_baselines}). No baseline wins a round against IG (0 of 48), and both judges rank all four below every agent-based system except Single-Agent Coding.

\begin{table}[H]
\centering
\caption{Statistical baselines judged against the fold-1 report pool (the first of three report replicates per configuration on SSB and HLE). Win rate (\%) is averaged over the Claude Opus~4.6 and GPT-5.4 judges; rubric is the 0--3 composite against each benchmark's frozen union gold (Opus~4.6). Cost is per report.}
\label{tab:stat_baselines}
\footnotesize\setlength{\tabcolsep}{3pt}
\begin{tabular}{l l l ccc cc}
\toprule
& & & \multicolumn{3}{c}{Win rate (\%)} & \multicolumn{2}{c}{Rubric} \\
\cmidrule(lr){4-6} \cmidrule(lr){7-8}
System & LLM use & Cost & SSB & HLE & APEX & SSB & APEX \\
\midrule
IG & Multi-agent & \$17 / \$71 / \$174 & \textbf{78.6} & \textbf{85.7} & \textbf{94.2} & \textbf{1.33} & \textbf{1.65} \\
CC Subagents & Multi-agent & \$38 & 75.0 & 78.6 & 88.5 & 1.32 & 1.26 \\
(4) Statistics + LLM write-up & One call & \$0.13--0.25 & 23.2 & 19.6 & 40.4 & 0.62 & 0.56 \\
(3) Retrieved traces + LLM write-up & One call & \$0.23--0.28 & 14.3 & 16.1 & 26.9 & 0.50 & 0.59 \\
(2) Combined statistics & None & \$0.04--0.11 & 7.1 & 7.1 & 3.8 & 0.09 & 0.25 \\
(1) Liblit & None & Negligible & 0.0 & 0.0 & 3.8 & 0.02 & 0.09 \\
\bottomrule
\end{tabular}
\end{table}

\subsection{Open-Weight Backbones}
\label{app:open_models}
We ran IG and the four comparison systems on two open-weight models with a 1M-token context, DeepSeek V4 Flash and GLM-5.3 Flash, with one report per system and benchmark. Each report was judged pairwise against the other systems on its backbone and the four statistical baselines, under both judges (Table~\ref{tab:open_models}). IG ranks first or ties for first in 7 of 9 backbone-benchmark cells, and against the Opus~4.6 pool IG on GLM-5.3 Flash wins every pairwise game under both judges. Downstream, the SSB patcher loop with IG reports reaches 0.828 (GLM-5.3 Flash) and 0.798 (DeepSeek V4 Flash) at the validation-selected round, against 0.80 and 0.879 in the two Opus~4.6 trials. On APEX-Agents, IG reports from GLM-5.3 Flash match Opus~4.6 at the validation-selected round (28 of 144 for both) and reach 34 at the best test round against 42, while DeepSeek V4 Flash reports reach 17 and 27.
An IG report costs \$8, \$18, and \$19 on SSB, HLE, and APEX-Agents with GLM-5.3 Flash and \$5, \$12, and \$17 with DeepSeek V4 Flash, against \$17, \$71, and \$174 with Opus~4.6, estimated from logged tokens at list prices. We first tried Qwen and Llama, but neither could run IG: the open release of Qwen3.8-2.4T-A95B has a 262k-token context, which IG's Orchestrator exceeded before submitting a report, and Llama 4 Maverick does not call tools reliably.

\begin{table}[H]
\centering
\caption{Open-weight backbones: pairwise win rate (\%) within each backbone's pool (the five systems on that backbone and the four statistical baselines, 16 games each), averaged over the Claude Opus~4.6 and GPT-5.4 judges. IG and the two strongest baselines per backbone, sorted by the average over the three benchmarks.}
\label{tab:open_models}
\small
\begin{tabular}{l l cccc}
\toprule
Backbone & System & SSB & HLE & APEX & Avg \\
\midrule
GLM-5.3 Flash & IG & 97 & 94 & 97 & 96.0 \\
Opus 4.6 & IG & 84 & 91 & 91 & 88.7 \\
DeepSeek V4 Flash & IG & 91 & 94 & 81 & 88.7 \\
Opus 4.6 & CC Subagents & 88 & 88 & 88 & 88.0 \\
GLM-5.3 Flash & Trace2Skill & 78 & 88 & 78 & 81.3 \\
DeepSeek V4 Flash & Trace2Skill & 84 & 78 & 78 & 80.0 \\
GLM-5.3 Flash & CC Subagents & 81 & 75 & 84 & 80.0 \\
DeepSeek V4 Flash & CC Subagents & 63 & 84 & 91 & 79.3 \\
Opus 4.6 & Trace2Skill & 72 & 84 & 75 & 77.0 \\
\bottomrule
\end{tabular}
\end{table}

\subsection{Filtering Unsupported Findings}
\label{app:filter}
An LLM entailment check (Claude Opus~4.6) reads up to three of a finding's cited traces with their outcomes and rates the finding as supported, partly supported, or not supported. Across all systems' reports, 81.5\% of the 254 SSB findings and 82.0\% of the 278 HLE findings are fully supported. We applied the check to IG's APEX-Agents reports at each budget tier of Table~\ref{tab:budget_scaling}, removed every finding rated not supported, and re-judged the filtered reports against the 10-system APEX-Agents pool (Table~\ref{tab:filter}). Removal lowers the judged win rate at every tier, and the judges prefer the unfiltered report in 13 of 16 head-to-head rounds, which suggests the pairwise judges reward the number of findings even when some are unsupported.

\begin{table}[H]
\centering
\caption{Filtering unsupported findings from IG's APEX-Agents reports. Win rate (\%) against the 10-system pool, averaged over the Claude Opus~4.6 and GPT-5.4 judges; the uncapped report is the pool's own IG report, so its row is over the other nine systems.}
\label{tab:filter}
\small
\begin{tabular}{l c c c c}
\toprule
IG budget cap & Findings & Not supported (removed) & Win rate, unfiltered & Win rate, filtered \\
\midrule
\$10 & 7 & 1 & 22.5 & 12.5 \\
\$25 & 9 & 2 & 42.5 & 20.0 \\
\$50 & 9 & 3 & 52.5 & 45.0 \\
Uncapped & 10 & 1 & 91.7 & 83.3 \\
\bottomrule
\end{tabular}
\end{table}

\subsection{Human Participant Study Protocol}
\label{app:participant_protocol}

We conducted three human-expert studies. Experiment~1 is the Expert Rating Study; Experiments~2 and~3 form the Expert Intervention Study on SSB and APEX-Agents. Experiment~1
(Section~\ref{sec:he_judge}) elicits expert ratings of individual
insights drawn from competing reports against a structured rubric,
isolating report quality from downstream implementation. Experiment~2
(Section~\ref{sec:he_intervention}) measures whether IG reports help
practitioners make better scaffold modifications under a held-out
benchmark protocol. Experiments~1 and~2 use SSB
\citep{spreadsheetbench} as the underlying corpus and CC Subagents as the comparison report source, and both blind participants
to the system that produced each report they evaluated.

\paragraph{Recruitment and eligibility.} Participants were professional
software engineers with experience in LLM agent development, recruited
through professional networks. Each participant had to be capable of reading agent
traces, diagnosing failure modes, and reasoning about scaffold code in
a real development environment. 

\subsubsection{Experiment 1: Expert Rating Study}
\label{app:exp1_protocol}

\paragraph{Materials.} Each participant received (i) the
SSB training-split trace corpus (250 agent traces: 110
correct, 140 incorrect; baseline pass rate 44\%); (ii) a benchmark
description covering the agent's code-execution loop, the available
tooling (\texttt{openpyxl} and \texttt{pandas}), the 5-turn budget,
and the cell-by-cell evaluation procedure; and (iii) a sequence of
insights drawn from the two reports under comparison. Insights were
presented one at a time and the source system was not disclosed. The
trace corpus was provided as a single bundled file
($\sim$6{,}000~lines); participants were instructed to use a coding
assistant (Claude Code or Cursor) to navigate and spot-check claims
rather than read traces sequentially.

\paragraph{Insight presentation.} Each insight followed a fixed
structured format identical across both report systems: a status
label (\texttt{confirmed}, \texttt{refuted}, or \texttt{inconclusive}),
a corpus-level prevalence estimate, a narrative description of the
pattern, an evidence block with cited trace IDs and verbatim quotes
each accompanied by a per-trace explanation, the full list of
affected trace IDs, a suggested scaffold action, and any additional
observations. The shared format prevents presentation-style confounds
between systems.

\paragraph{Rubric.} Each insight was scored on a 1--5 scale across
four dimensions: \emph{correctness} (whether the claim is actually
true against the traces), \emph{depth} (whether the insight explains
the underlying mechanism rather than describing the symptom),
\emph{evidence quality} (whether the claim is backed by specific,
verifiable trace citations), and \emph{actionability} (whether the
suggested improvement is concrete, proportionate to the evidence, and
free of overfitting risk). Anchor descriptions were provided for each
level on each dimension; per-dimension score
breakdowns are reported in
Appendix~\ref{app:he_judge_details}. Rubric instructions explicitly
flagged common confounds: \emph{correctness} and \emph{evidence
quality} are independent (a finding may cite many traces but draw
the wrong conclusion, or be true and verifiable without citing
specifics); a finding that claims population-level prevalence while
citing only 1--2 traces was capped at evidence-quality $\leq 3$; and
a finding that recommended a sweeping change based on a narrow
pattern received a lower actionability score regardless of its
correctness on the cited traces.

\paragraph{Time budget and grading discipline.} Approximately 120
minutes total per participant, with a 5--10~minute target per
insight. Participants recorded a start time so that per-insight
grading time could be measured post-hoc. Participants were instructed
to score each insight on its own merits, not to adjust scores based
on how other insights had scored, and not to deflate scores when two
insights described overlapping phenomena.

\paragraph{Cohort and sample.} Twelve human-expert evaluators scored
all 23 insights (11 IG, 12 CC Subagents) of one IG report and one CC Subagents report generated on the SSB
training split. Each (insight,~evaluator) pair produced one set of
four dimension scores; aggregate results in Section~\ref{sec:he_judge}
are means over evaluators within each report condition.

\subsubsection{Experiment 2: Expert Intervention Study on SSB}
\label{app:exp2_protocol}

\paragraph{Random assignment and blinding.}
Eligible participants were randomly assigned to one of the two report
conditions (IG or CC Subagents). Assignment labels were not disclosed:
participants were told that they would be evaluating one of several
candidate analysis-report formats but were not informed which system had
generated the report they received, and the report itself contained no
system-identifying metadata. Graders of the resulting scaffold
modifications had access only to the held-out benchmark task set and the
participant's modified scaffold; they were not told which report condition
the participant had been in.

\paragraph{Baseline-competency screen.}
Prior to the main intervention task, each participant completed a short
qualification task designed to confirm baseline competence with the trace
data and tooling: given a sample of execution traces, the participant was
asked to use a coding assistant (Claude Code or Cursor) to compute the
distribution of tool calls per trace. Participants who completed the
screen task were enrolled in the main study; the screen was scored
pass/fail on whether the reported counts matched ground truth on a small
held-out sample. The screen was identical across conditions and was
completed before condition assignment was revealed, so it cannot bias the
between-arm contrast.

\paragraph{Conditions.} Two conditions: participants received either
an IG report or a CC Subagents report (six participants per
arm). Condition assignment was blinded so participants did not know
which report-generation system produced the analysis they received.
Headline results in Section~\ref{sec:he_intervention} compare both
arms against the unmodified baseline scaffold.

\paragraph{Task briefing.} Each participant received a self-contained
package ($\sim$400~MB) containing: (i) the SSB scaffold
source; (ii) 250 training traces and 50 baseline-eval traces; (iii)
the assigned condition's insight report; (iv) the reference baseline
pass rate (44\% on 22/50 dev tasks); (v) a Docker runtime image with
Python, LibreOffice, the benchmark eval data, and the harness; and
(vi) a Claude Code warm-up task to verify environment setup.
Participants were instructed to use Claude Code or Cursor as their
coding assistant and were provided a \$50 API budget for agent calls
(Claude Opus 4.6).

\paragraph{Train/eval split.} The 250 training traces were generated
with GPT-5.2 and serve as the failure-mode corpus that each insight
report describes; the 50-task evaluation set is run with Claude Opus
4.6, which is also the model the participant's modified scaffold is
scored against. Train and eval task sets are disjoint, and because training traces
come from GPT-5.2 while evaluation runs Claude Opus~4.6, baseline pass
rates are not comparable between the two. Participants were instructed to
treat training traces as illustrative of scaffold failure modes
rather than as a performance ceiling.

\paragraph{Time and iteration.} Maximum 1.5 hours per participant.
Iterative scaffold edits via the participant's IDE were picked up
automatically by the running container, allowing rapid edit/eval
cycles.

\paragraph{Locked configuration.} Four scaffold parameters were
locked and the eval runner enforced the lock: the model (Claude Opus
4.6), the maximum number of agent turns (5), the number of preview
rows shown to the agent (5), and the per-cell execution timeout
(60\,s). Participants modified the prompts, the agent loop, and
non-locked configuration fields only. Modifying the eval runner, the
task-id list, or the data directory invalidated a submission.

\paragraph{Evaluation.} Each participant's modified scaffold was
scored on the 100-task held-out test split (disjoint from the 250
training tasks and the 50-task dev set used during iteration). Scoring was performed by re-running the eval
harness three times per submission and averaging; reported scores in
Section~\ref{sec:he_intervention} are these per-participant means.
Evaluation is deterministic per run (cell-by-cell spreadsheet
comparison), so run-to-run variance reflects only the agent's
non-determinism at Claude Opus 4.6.

\paragraph{Submission and audit trail.} Participants exported their
coding-assistant chat transcripts as part of the submission, ordered
chronologically. The transcript audit trail was used post-hoc to
verify that participants did not bypass the locked configuration or
access eval-side artifacts. No participant in either report condition
was excluded for protocol violations.

\paragraph{Compensation and labor.} To maintain study quality,
we provided participants in Experiments~1 and~2 with a
comprehensive briefing package governing study procedures,
evaluation rubrics, and (for Experiment~2) the
scaffold-improvement task and locked-configuration constraints.
Due to the inclusion of personally identifying recruitment
correspondence and proprietary internal protocols, we summarize
the core requirements above and do not release the unredacted
briefing materials. On average, each Experiment~1 grading
session required approximately 2~hours of expert labor to
review traces, score insights, and submit responses; each
Experiment~2 intervention session required up to 1.5~hours of
expert labor to analyze the assigned report, modify the
scaffold, and submit results. Participants in Experiments~1, 2, and~3
were compensated under contractual terms that met or exceeded
applicable local wage requirements. No unpaid volunteer or
crowd labor was used.

\subsubsection{Experiment 3: Expert Intervention Study on APEX-Agents}
Experiment~3 repeats the Experiment~2 design on APEX-Agents: random
assignment to the IG or CC Subagents report condition, blinding to
report source, identical tooling and benchmark access, and a fixed
two-hour budget. Participants iterated against a 20-task development
split whose evaluation takes up to ten minutes per run, and were scored
on the 144-task held-out split.

\subsection{Statistical Analysis of the Expert Intervention Study}
\label{app:stats}

This appendix provides robustness analyses supporting the Welch two-sample
$t$-test reported in Section~\ref{sec:he_intervention}. The SSB analyses below operate
on the per-participant data ($n=6$ per arm; each participant's score is the
mean of three independent evaluation runs).

\paragraph{Welch two-sample $t$-test (primary).}
The CC Subagents arm has a per-participant mean of $43.2\%$ (SD $9.95$,
SEM $4.06$); the IG arm has $57.4\%$ (SD $3.69$, SEM $1.51$). The
between-arm difference is $14.17$\,pp with standard error $4.33$, yielding
Welch's $t = 3.27$, Welch--Satterthwaite $\mathrm{df} = 6.35$, and
two-sided $p = 0.016$. The 95\% confidence interval on the between-arm
difference is $[3.56, 24.78]$\,pp; Cohen's $d = 1.89$.

\paragraph{Exact permutation test.}
As a non-parametric robustness check that does not assume normality of the
per-participant means, we enumerate all $\binom{12}{6} = 924$ ways of
relabeling the 12 participants between arms. The observed difference
($14.17$\,pp) is matched or exceeded in $12$ of $924$ relabelings, giving an
exact two-sided $p = 0.013$, in agreement with the parametric Welch result.

\paragraph{Leave-one-participant-out sensitivity.}
We refit the Welch test 12 times, each time omitting one of the 12
participants. The between-arm difference ranges from $12.4$ to $17.9$\,pp;
the test statistic ranges $t \in [2.60, 6.96]$ and the two-sided $p$-value
ranges $p \in [0.0001, 0.048]$, with $p < 0.05$ in every refit. The
worst-case refit leaves $p \approx 0.048$; the best-case
refit leaves $p < 0.001$. No single participant is responsible
for the headline result.

\paragraph{Mixed-effects model.}
Because each participant contributes three evaluation runs, we additionally
fit a random-intercepts linear mixed model to the
$2 \times 6 \times 3 = 36$ run-level scores using REML, with arm as a fixed
effect and participant as a random intercept:
\[
y_{ijk} = \beta_0 + \beta_1 \cdot \mathbf{1}[\mathrm{arm}_{ij} = \text{IG}] + u_j + \epsilon_{ijk},
\quad u_j \sim \mathcal{N}(0, \sigma_u^2),\ \epsilon_{ijk} \sim \mathcal{N}(0, \sigma_\epsilon^2).
\]
The fitted arm contrast is $\hat{\beta}_1 = 14.17$~pp ($\mathrm{SE} = 4.33$,
$z = 3.27$, $p = 0.001$; with 12 participants a normal reference is anti-conservative, so we rely on the Welch result). The estimated participant random-effect variance
is $\hat{\sigma}_u^2 = 53.3$ and the residual variance is
$\hat{\sigma}_\epsilon^2 = 8.8$, giving an intraclass correlation of
$\mathrm{ICC} = 0.86$. The high ICC indicates that nearly all of the
score variance is between-participant rather than within-participant, which
justifies aggregating to per-participant means in the primary analysis.

\paragraph{Sensitivity (minimum detectable effect).}
A two-sample $t$-test with $n_1 = n_2 = 6$ at $\alpha = 0.05$ (two-sided)
achieves power $\geq 0.80$ for standardized effect sizes
$d \gtrsim 1.8$. We do not claim power to detect smaller effects; a
larger replication remains an important direction for future work.

\paragraph{APEX-Agents intervention study.}
Sixteen participants (8 per arm) completed the SSB
design on APEX-Agents (236-trace corpus, 20 development tasks, 144
held-out test tasks, two-hour budget). Per-arm means are 20.49\% (SE 0.63\,pp) for IG and 18.40\% (SE 0.75\,pp) for CC Subagents, against an
unmodified-scaffold baseline of 19.44\%. Table~\ref{tab:apex_stats}
reports the tests on the 16 per-participant means. At $n=8$ per arm a
two-sample $t$-test at $\alpha=0.05$ has power $\geq 0.80$ only for
standardized effects $d \gtrsim 1.5$, so the study was underpowered for the observed $d = 1.06$. The percentile bootstrap interval excludes zero, but at $n=8$ per arm it is less conservative than the three tests, none of which reaches $p<0.05$.

\begin{table}[H]
\centering
\caption{Significance tests for the APEX-Agents Expert Intervention Study, IG vs.\ CC Subagents on the 16 per-participant means ($n=8$ per arm). Bootstrap interval in percentage points (pp).}
\label{tab:apex_stats}
\small
\begin{tabular}{lll}
\toprule
Test & Statistic & $p$ \\
\midrule
Two-sided Welch $t$ & $t = 2.121$ & 0.053 \\
Mann--Whitney $U$ (exact) & $U = 14$ & 0.065 \\
Exact permutation (all 12{,}870 label assignments) & -- & 0.066 \\
Bootstrap 95\% CI of the difference & $[+0.26, +3.82]$\,pp & -- \\
Cohen's $d$ & 1.06 & -- \\
\bottomrule
\end{tabular}
\end{table}

\subsection{Expert Rating Study: Per-Dimension Scores}
\label{app:he_judge_details}

\begin{table}[H]
\centering
\caption{Expert Rating Study on SSB: mean 1--5 ratings by 12 evaluators over every insight of one IG report ($N=11$) and one CC Subagents report ($N=12$), per dimension and averaged (Avg). Parentheses: SEM across insights, after averaging each insight over the 12 evaluators.}
\label{tab:he_judge}
\footnotesize\setlength{\tabcolsep}{3.5pt}
\begin{tabular}{lcccccc}
\toprule
Report & $N$ & Correctness & Depth & Evidence & Actionability & Avg \\
\midrule
CC Subagents & 12 & 4.278 ($\pm$0.08) & 4.139 ($\pm$0.09) & 4.285 ($\pm$0.08) & 4.167 ($\pm$0.12) & 4.217 ($\pm$0.07) \\
IG & 11 & 4.136 ($\pm$0.12) & 4.311 ($\pm$0.10) & 4.341 ($\pm$0.12) & 4.220 ($\pm$0.11) & 4.252 ($\pm$0.10) \\
\bottomrule
\end{tabular}
\end{table}

\subsection{Automated Patcher Loop: Protocol and Per-Round Detail}
\label{app:patcher_loop}

\subsubsection{Protocol}
With $\mathcal{A}_0$ as the base scaffold, at round~$r$, traces
$\mathcal{T}_r$ are collected by executing scaffold $\mathcal{A}_r$
against the benchmark training split. The analysis system under test
produces a report $R_r$ from $\mathcal{T}_r$, which the patcher (Claude
Code with Claude Opus~4.6 as the underlying model) consumes alongside the
scaffold source code to produce $\mathcal{A}_{r+1}$. Each round's
modified scaffold is evaluated on a validation split; the loop
terminates when validation pass rate fails to improve by
$\varepsilon=0.01$ for two consecutive rounds, or after five rounds on SSB in trial 1 and eight in trial 2 (APEX-Agents settings below),
whichever comes first. We report held-out test pass rate at each round, and summarize each arm by its test score at the best validation round and its best test score. The second SSB trial repeats every arm on the same split with an eight-round cap.

\subsubsection{Conditions}
In every trial we compare four conditions: (i) the IG report; (ii) RLM~\citep{rlm};
(iii) CC Subagents (the same analysis system used in
Sections~\ref{sec:he_intervention} and~\ref{sec:he_judge}); and (iv) a
pure patcher, in which the patcher receives
only the scaffold source code, isolating what the patcher achieves
without any analysis input; the second SSB trial also adds a Trace2Skill report and the raw training traces with no report.

\subsubsection{Best held-out scores on SSB and APEX-Agents}
Table~\ref{tab:patcher_full} gives each arm's validation-selected and best test score in every trial; the trial-1 SSB per-round test and validation trajectories follow (Table~\ref{tab:patcher_loop_appendix}), and the APEX-Agents budget grid, component ablations, and budget scaling are in the APEX-Agents subsection below.

\begin{table}[H]
\centering
\caption{Automated patcher loop: held-out score per arm at the validation-selected round (Val) and the best test round (Best). SSB: pass rate, trial 1 with a five-round cap and trial 2 with an eight-round cap. APEX-Agents: uncapped analysis budget, tasks solved of 144, where validation predicts test only weakly ($r = 0.25$). Raw traces: the patcher reads the training traces with no report. Unmodified SSB: round-0 pass rate (0.41--0.44 across arms, Table~\ref{tab:patcher_loop_appendix}); unmodified APEX-Agents scaffold is pooled over five trials. --: not run.}
\label{tab:patcher_full}
\small
\begin{tabular}{l cc cc cc c}
\toprule
& \multicolumn{2}{c}{SSB, trial 1} & \multicolumn{2}{c}{SSB, trial 2} & \multicolumn{2}{c}{APEX-Agents} & APEX \\
\cmidrule(lr){2-3} \cmidrule(lr){4-5} \cmidrule(lr){6-7}
Arm & Val & Best & Val & Best & Val & Best & rounds \\
\midrule
IG & 0.80 & 0.84 & 0.879 & 0.879 & 28 & 42 & 7 \\
CC Subagents & 0.76 & 0.81 & 0.818 & 0.838 & 23 & 32 & 4 \\
RLM & 0.79 & 0.84 & 0.818 & 0.838 & 30 & 32 & 4 \\
Trace2Skill & -- & -- & 0.808 & 0.828 & -- & -- & -- \\
Raw traces, no report & -- & -- & 0.828 & 0.828 & 39 & 39 & 2 \\
Pure patcher (no traces, no report) & 0.80 & 0.80 & 0.758 & 0.758 & 35 & 35 & 3 \\
Unmodified scaffold & 0.41--0.44 & 0.41--0.44 & -- & -- & 30.8 & 30.8 & -- \\
\bottomrule
\end{tabular}
\end{table}

\subsubsection{Per-round trajectories}

\begin{table}[H]
\centering
\caption{Patcher-loop trajectory on SSB (trial 1): held-out test and validation pass rate of scaffold $\mathcal{A}_r$ after round $r$. The validation series drives the patience stopping rule. ``--'': loop stopped earlier on the validation criterion. Bold: row-wise peak.}
\label{tab:patcher_loop_appendix}
\small
\begin{tabular}{lcccccc}
\toprule
Method & $\mathcal{A}_0$ & $\mathcal{A}_1$ & $\mathcal{A}_2$ & $\mathcal{A}_3$ & $\mathcal{A}_4$ & $\mathcal{A}_5$ \\
\midrule
\multicolumn{7}{l}{\textit{Test pass rate}} \\
IG            & 0.41 & 0.77 & 0.81 & 0.80 & \textbf{0.84} & 0.80 \\
RLM           & 0.42 & 0.79 & 0.81 & \textbf{0.84} & --   & --   \\
CC Subagents  & 0.44 & 0.76 & 0.71 & \textbf{0.81} & --   & --   \\
Pure patcher  & 0.41 & \textbf{0.80} & 0.58 & 0.58 & --   & --   \\
\midrule
\multicolumn{7}{l}{\textit{Validation pass rate}} \\
IG            & 0.46 & 0.78 & 0.84 & \textbf{0.86} & \textbf{0.86} & 0.84 \\
RLM           & 0.22 & \textbf{0.82} & 0.80 & \textbf{0.82} & --   & --   \\
CC Subagents  & 0.24 & \textbf{0.86} & 0.68 & 0.78 & --   & --   \\
Pure patcher  & 0.42 & \textbf{0.72} & 0.56 & 0.48 & --   & --   \\
\bottomrule
\end{tabular}
\end{table}

\subsubsection{IG report content evolves across rounds}
The findings IG introduces at successive rounds describe failure modes at
progressively finer levels of granularity. At round~1, the dominant
finding identifies the agent writing Excel formula strings into cells
using \texttt{openpyxl}, which stores them as literal text rather than
evaluating them, a pattern accounting for 84\% of round-1 failures. At
round~3, the report identifies a scaffold control-flow bug in which a
fallback path executes the model's natural-language response as Python
when no fenced \texttt{python} block is present. At round~5, the report
identifies an \texttt{openpyxl} API version-skew (the agent uses
\texttt{wb.defined\_names.definedName}, removed in the installed
version), and a save-and-reload sequence that shifts row indices and
breaks formula references in cells written earlier in the same task.
Across rounds, IG's report length grows from 4{,}074 to 4{,}816 words
and the number of distinct findings grows from 7 to 9.

\subsubsection{Pure patcher round-2 reversal}
The pure-patcher condition reaches the highest round-1 test score in the
cohort (0.80) and then regresses to 0.58 at round~2 and again at round~3.
We trace the round-2 regression to a single round-2 prompt edit: the
round-2 patcher, lacking a regenerated analysis input, added the
instruction ``If you are unsure about a value, write your best estimate
rather than leaving it as None --- an approximate value is better than a
missing one.'' SSB grades cells by exact-value match,
including expected \texttt{None} values for legitimately blank answer
cells; the round-1 scaffold passed those tasks by leaving the cells
empty. The round-2 instruction directs the agent to fill them with
heuristic guesses, and 12 of the 13 round-2 regressions exhibit the eval
detail \texttt{expected None, got <sentinel>} (where the sentinel is a
whitespace character, the string \texttt{"N/A"}, or a hallucinated
value). Without an analysis input grounded in the round-1 traces, the
patcher's round-2 priorities are derived from the scaffold source and
its own priors, with no benchmark-grounded check; the resulting edit is
confidently structural but contradicts the eval contract.

\subsubsection{APEX-Agents loop, budget grid, and component ablations}
\label{app:apex_loop}
The APEX-Agents runs use the loop above with the analysis system's per-round LLM spend capped at \$10, \$25, or \$50, or left uncapped; realized spend is approximate and occasionally deviates from the nominal cap. The loop stops after two rounds without a validation improvement of at least 0.01, up to a maximum of eight rounds (the SSB loop was capped at five rounds in trial 1 and eight in trial 2). At the \$10 and \$25 caps the per-round budget was exhausted before any Investigator was dispatched, so those reports consist of Scout and Orchestrator findings without Investigator validation; the confirmed-only rule of Algorithm~\ref{alg:ig_loop} applies whenever the budget permits Investigator dispatch. Table~\ref{tab:apex_loop} reports the best held-out test score per arm, Table~\ref{tab:ablation_judge} the component ablations at the \$50 cap with their judged report quality, and Table~\ref{tab:budget_scaling} the within-IG budget scaling. In the ablation-pool tournament each arm plays eight games and no ties were recorded. The IG loop also recovered from a round-1 regression in which the patcher added a planning step that made the agent emit plans instead of answers: the next IG report identified the failure and the patcher reverted it, whereas the pure patcher did not recover from its breaking change.

\paragraph{Statistical caveats.} The unmodified-scaffold baseline is pooled over five independent trials (154/720 tasks, 0.214, SD 2.59 tasks), and each arm's significance against it is a two-sided prediction-interval $t$-test that treats the arm's single trial as one new draw from the baseline trial distribution (standard error $s\sqrt{1+1/5}$, four degrees of freedom). The reported score is the best held-out result across the loop's rounds and the baseline variance does not include patcher stochasticity, so these $p$-values are optimistic. IG ($p = 0.017$) is significant at the unadjusted level but not after a Holm correction across the four uncapped arms. Its ten-task margin over the CC Subagents and RLM reports is a difference between two single trials (standard error $2.59\sqrt{2} \approx 3.7$ tasks, $t \approx 2.7$ on four degrees of freedom, two-sided $p \approx 0.05$). The IG loop completed seven rounds before the patience rule stopped it, against four for CC Subagents and RLM, which gives it more rounds from which the best score is taken. Across the capped grid every cell lies within 27--35/144, the best capped cell (RLM at \$25, 35/144) is 1.6 baseline standard deviations above the pooled baseline, and neither CC Subagents nor RLM improves monotonically as the cap rises; realized spend was not matched, since RLM stayed near \$12 per round at every cap whereas CC Subagents spent up to its cap. Among the \$50-cap ablations, only the no-LLM-extraction arm (19/144) lies outside single-trial baseline noise; the other removals cost two to four tasks, within it.

\begin{table}[H]
\centering
\caption{APEX-Agents automated patcher loop: best held-out test score (tasks solved of 144; Pass@1 in parentheses) per analysis system and per-round analysis cap. Unmodified scaffold: 30.8/144 (0.214), pooled over five trials, SD 2.59 tasks. $^{*}$Above the baseline prediction interval at $p<0.05$ (two-sided; unadjusted). Bold: best cell.}
\label{tab:apex_loop}
\small
\begin{tabular}{l cccc}
\toprule
Analysis system & \$10 cap & \$25 cap & \$50 cap & Uncapped \\
\midrule
IG & 33 (0.229) & 34 (0.236) & 34 (0.236) & \textbf{42 (0.292)}$^{*}$ \\
CC Subagents & 27 (0.188) & 33 (0.229) & 27 (0.188) & 32 (0.222) \\
RLM & 34 (0.236) & 35 (0.243) & 32 (0.222) & 32 (0.222) \\
Pure patcher (no traces, no report) & -- & -- & -- & 35 (0.243) \\
\bottomrule
\end{tabular}
\end{table}

\begin{table}[H]
\centering
\caption{IG component ablations. SSB patcher: held-out pass rate at the validation-selected (Val) and best test (Best) round in the second patcher trial; the no-Python-REPL arm was not run on SSB. APEX patcher: best held-out test score (of 144) at a \$50 per-round analysis cap. WR and Comp.\ are on APEX-Agents. WR: pairwise win rate (\%) within the five-report ablation pool only (full IG and the four ablations), so not comparable to Table~\ref{tab:llm_judge_combined}; Comp.: rubric composite (0--3) against the union-gold set of the 10-system APEX judge run (Table~\ref{tab:cross_judge}). Both are shown under the Opus~4.6 and GPT-5.4 judges and score each arm's uncapped report rather than its \$50-cap run. Comp.\ does not penalize citations of trace IDs absent from the corpus (18.2\% of the no-LLM-extraction report's findings, 0\% for full IG), whereas WR and Patcher do.}
\label{tab:ablation_judge}
\small\setlength{\tabcolsep}{4pt}
\begin{tabular}{l cc c cc cc}
\toprule
& \multicolumn{2}{c}{SSB patcher} & APEX & \multicolumn{2}{c}{WR (\%), ablation pool} & \multicolumn{2}{c}{Comp.\ (0--3)} \\
\cmidrule(lr){2-3} \cmidrule(lr){5-6} \cmidrule(lr){7-8}
Arm & Val & Best & patcher & Opus~4.6 & GPT-5.4 & Opus~4.6 & GPT-5.4 \\
\midrule
Full IG & 0.879 & 0.879 & 34 & 87.5 & 100.0 & 1.50 & 1.49 \\
Orchestrator-only (no dispatch) & 0.828 & 0.848 & 32 & 25.0 & 37.5 & 0.92 & 0.95 \\
Generic Subagents (no typing) & 0.838 & 0.838 & 31 & 75.0 & 75.0 & 1.47 & 1.38 \\
No Python REPL & -- & -- & 30 & 0.0 & 12.5 & 1.03 & 1.23 \\
No LLM extraction & 0.838 & 0.859 & 19 & 62.5 & 25.0 & 1.68 & 1.76 \\
\midrule
Unmodified scaffold & -- & -- & 30.8 & \multicolumn{4}{c}{} \\
\bottomrule
\end{tabular}
\end{table}

\begin{table}[H]
\centering
\caption{IG budget scaling on the APEX-Agents patcher loop. Val: held-out score at the validation-selected round; Best test: best held-out score, an upper bound (both of 144). WR: pairwise win rate (\%) within a five-report pool of IG reports (the four tiers shown plus a \$100-cap report, not shown), so not comparable to Table~\ref{tab:llm_judge_combined}; Comp.: rubric composite (0--3) from the same judge run over the tier reports, which builds its own union-gold set and is therefore not comparable to Table~\ref{tab:cross_judge}. WR and Comp.\ are shown under the Opus~4.6 / GPT-5.4 judges. Scouts, Investigators (Invest.), and Orchestrator (Orch.) turns are per-round means; Cite valid.: fraction of cited trace IDs that resolve; count: citations per report.}
\label{tab:budget_scaling}
\scriptsize\setlength{\tabcolsep}{3pt}
\begin{tabular}{l c c c c c c c c c}
\toprule
& & & & \multicolumn{1}{c}{WR (\%)} & \multicolumn{1}{c}{Comp.} & & & & \\
IG cap & Rounds & Val & Best test & Opus / GPT & Opus / GPT & Scouts & Invest. & Cite valid., count & Orch.\ turns \\
\midrule
\$10 & 5 & 33 & 33 & 12.5 / 0.0 & 1.24 / 1.24 & 1 & 0 & 81.6\%, 111.8 & 16.5 \\
\$25 & 4 & 27 & 34 & 25.0 / 25.0 & 1.36 / 1.44 & 5 & 0 & 84.3\%, 129 & 16.8 \\
\$50 & 4 & 32 & 34 & 50.0 / 62.5 & 1.61 / 1.86 & 8.67 & 1.67 & 95.7\%, 256.7 & 19.3 \\
Uncapped (\$174/round) & 7 & 28 & 42 & 87.5 / 75.0 & 1.87 / 1.83 & 8.33 & 12 & 99.8\%, 633.3 & 35.2 \\
\bottomrule
\end{tabular}
\end{table}

\subsection{Patcher Prompt}
\label{app:patcher_prompt}

The Patcher prompt template is rendered at runtime with conditional
sections for the four input conditions (codebase only; codebase with
traces; codebase with report; codebase with report and traces); the
template below shows the form with both an IG report and the trace bundle
present. For the codebase-only condition the IG-report and traces sections
are dropped. The benchmark-specific specification is appended verbatim
after these instructions.

\begin{quote}\small\ttfamily\raggedright
You are a coding agent tasked with improving an agent scaffold to
maximize its benchmark performance.

\textbf{Your workspace.} The source directory contains the agent
scaffold source code --- this is what you modify. An IG report file
contains an analysis of the agent's execution traces with identified
failure modes and recommended fixes; start by reading it. A traces
directory contains the agent's execution traces; use them to understand
where the agent fails and why.

\textbf{Workflow.} Approach this rigorously --- plan first, then
implement. Premature edits produce shallow patches that don't
generalize. Most of your effort should be in steps 1--2.

(1) Explore. Get a thorough understanding of the scaffold code --- its
entry points, the tool surface, and where the agent's behavior is
configured vs hard-coded. Read the IG report carefully for the
failure-mode analysis and recommended interventions. Examine traces to
identify concrete failure patterns from real runs. You may dispatch
focused-exploration subagents and use a TodoWrite tool to track your
plan as you progress.

(2) Diagnose and plan. For each failure mode, name (a) the specific
scaffold layer that's wrong (system prompt, tool definition, control
flow, retry policy, response synthesis); (b) the mechanism --- why the
current code produces that failure (not just the surface symptom); and
(c) the smallest concrete change that fixes it.

(3) Implement. Apply the changes from your plan. Prefer small, targeted
edits over rewrites.

(4) Evaluate. Call the evaluate-and-submit tool to test your changes.
With one attempt available, make your best changes then submit. With
multiple attempts, you receive evaluation results after each call and
can iterate.

\textbf{Constraints.} Only modify files in the source directory. Do not
access anything outside it. Do not change the model or LLM
configuration. Do not use web search or web fetch. Do not access, read,
or inspect any evaluation code, test data, ground truth, or benchmark
task definitions. Do not attempt to read task metadata, answer keys, or
expected outputs from the environment object. Focus on scaffold changes:
system prompts, tool implementations, error handling, control flow,
response synthesis.
\end{quote}

\subsection{Cost and Runtime Detail}
\label{app:compute}

Table~\ref{tab:ig_cost_full} reports per-system LLM cost and wall-clock
for a single analysis run, averaged over $n{=}6$ runs
(3 replicates $\times$ 2 benchmarks: SSB and HLE). Costs are
imputed from token counts at Anthropic's published list price
(\texttt{claude-opus-4-6}: \$5/M input, \$25/M output;
\texttt{claude-haiku-4-5}: \$1/M input, \$5/M output) where the source
schema does not persist a dollar field; otherwise costs are taken
verbatim from the harness's billing telemetry. Wall-clock is the elapsed
time of the analysis run itself, from the same source records.

\begin{table}[H]
\centering
\caption{LLM cost and wall-clock per analysis run, mean over $n{=}6$ runs (three replicates each on SSB and HLE). $^{*}$: cost imputed from recorded token counts at Anthropic list prices (Claude Opus~4.6: \$5/M input, \$25/M output; Claude Haiku~4.5: \$1/M input, \$5/M output); other costs are recorded by the harness. IG's wide range spans SSB ($\sim$\$17 mean) and the longer-trace HLE ($\sim$\$71 mean).}
\label{tab:ig_cost_full}
\small
\begin{tabular}{lrrrrrr}
\toprule
& \multicolumn{3}{c}{\textbf{LLM cost (\$)}} & \multicolumn{3}{c}{\textbf{Wall-clock (min)}} \\
\cmidrule(lr){2-4}\cmidrule(lr){5-7}
\textbf{System} & \textbf{Mean} & \textbf{Median} & \textbf{Range}
                    & \textbf{Mean} & \textbf{Median} & \textbf{Range} \\
\midrule
Single-Agent Coding    & 5.13 & 5.01 & 2.68 -- 8.20 & 14.6 & 13.3 & 9.0 -- 22.8 \\
RLM$^{*}$              & 7.72 & 7.36 & 5.09 -- 9.96 & 19.5 & 17.5 & 10.6 -- 29.3 \\
Trace2Skill            & 26.43 & 25.63 & 22.22 -- 32.49 & 37.3 & 35.8 & 28.6 -- 53.4 \\
CC Subagents          & 38.09 & 38.05 & 33.00 -- 42.95 & 34.4 & 34.3 & 27.2 -- 42.1 \\
IG$^{*}$        & 44.21 & 42.19 & 12.17 -- 81.55 & 48.4 & 48.6 & 38.1 -- 56.4 \\
\bottomrule
\end{tabular}
\end{table}

\end{document}